\documentclass{article}

\usepackage[preprint]{neurips_2026}

\usepackage[utf8]{inputenc} 
\usepackage[T1]{fontenc}    
\usepackage{hyperref}       
\usepackage{url}            
\usepackage{booktabs}       
\usepackage{amsfonts}       
\usepackage{nicefrac}       
\usepackage{microtype}      
\usepackage{xcolor}         
\usepackage{microtype}
\usepackage{hyperref}
\usepackage{url}
\usepackage{amssymb} 
\usepackage{booktabs}
\newcommand{\myparatight}[1]{\smallskip\noindent{\bf {#1}:}~}
\newcommand{\alg}{\textsf{CoRe-Code}\xspace}
\usepackage{bm}
\usepackage{xspace} 
\usepackage[table]{xcolor} 

\usepackage{subfig}
\usepackage{enumitem}
\usepackage{subcaption}  
\usepackage{caption}
\usepackage{hyperref}
\usepackage{wrapfig}
\usepackage{amsfonts}

\usepackage{multirow}
\usepackage{svg}
\definecolor{greyL}{RGB}{210,233,232}
\usepackage{algorithm}
\usepackage{algpseudocode}
\usepackage{hyperref}
\usepackage{url}
\usepackage{amsthm}
\usepackage{tcolorbox}
\tcbuselibrary{listings, breakable}
\usepackage{listings}
\usepackage{xcolor}
\definecolor{cvprblue}{rgb}{0.21,0.49,0.74}
\definecolor{greyL}{RGB}{230,248,255}
\definecolor{yellow}{RGB}{255,246,234}
\definecolor{green}{RGB}{230,247,239}
\definecolor{blues}{RGB}{226,234,249}
\usepackage{amsmath}
\usepackage{times}
\usepackage{latexsym}
\theoremstyle{plain}
\newtheorem{theorem}{Theorem}[section]

\theoremstyle{definition}
\newtheorem{definition}[theorem]{Definition}

\theoremstyle{remark}

\usepackage[T1]{fontenc}
\usepackage{times}
\usepackage{latexsym}

\usepackage[T1]{fontenc}

\usepackage[utf8]{inputenc}

\usepackage{microtype}

\usepackage{inconsolata}

\usepackage{graphicx}

\newtcolorbox{promptbox}[2][]{%
  breakable,
  before={\clearpage},
  after={\clearpage},
  colback=yellow,          
  colframe=blues,      
  coltitle=cvprblue,
  fonttitle=\bfseries,
  title=#2,
  boxrule=0.5pt,
  arc=2mm,
  outer arc=2mm,
  #1
}

\lstdefinestyle{mypython}{
  language=Python,
  basicstyle=\ttfamily\small,
  keywordstyle=\color{teal!70!black}\bfseries,
  commentstyle=\color{gray!70},
  stringstyle=\color{purple!70!black},
  breaklines=true,
  frame=single,
  framerule=0.3pt,
  rulecolor=\color{gray!40},
  backgroundcolor=\color{gray!5},
  columns=flexible
}

\lstdefinestyle{plantext}{
  language={},
  basicstyle=\ttfamily\small,
  breaklines=true,
  frame=single,
  framerule=0.3pt,
  rulecolor=\color{blue!20!black},
  backgroundcolor=\color{blue!2},
  columns=flexible
}

\usepackage{lineno}

\definecolor{darkblue}{rgb}{0, 0, 0.5}
\hypersetup{colorlinks=true, citecolor=darkblue, linkcolor=darkblue, urlcolor=darkblue}
\title{\alg: Collaborative Reinforcement Learning for Code Generation}

\author{
  Zhihao Dou\thanks{Equal contribution} \\
  Department of Computer Science\\
Case Western Reserve University\\
  \And
Qinjian Zhao\footnotemark[1] \\
  Department of Computer Science \\
  Kean University \\
  Union, NJ, USA \\
   \And
   Zhongwei Wan \\
  The Ohio State University \\
  Columbus, OH, USA \\
  \And
    Xiaoyu Xia \\
    Royal Melbourne Institute of Technology \\
    Melbourne, VIC, Australia \\
\And
Sumon Biswas \\
  Department of Computer Science \\
  Case Western Reserve University \\
  Cleveland, OH, USA \\
}

\begin{document}

\maketitle

\begin{abstract}
  Large language models (LLMs) have achieved strong performance in code generation, but most methods rely on autoregressive decoding without global planning, often leading to locally coherent yet globally suboptimal solutions (e.g., failing test cases or inefficient complexity). While recent approaches such as Chain-of-Thought (CoT) and multi-agent systems (MAS) introduce planning, their limited role specialization and coordination hinder performance on complex tasks.
To address the challenges of coordination and specialization in multi-agent code generation, we propose \textbf{Co}llaborative \textbf{Re}inforcement \textbf{Code} (\alg), a framework for role-specialized LLM agents that enhances inter-agent coordination to generate more accurate and efficient code. \alg adopts a simple Planner–Coder paradigm, where the Planner produces high-level plans and the Coder executes them to generate code. We further introduce a collaboration-aware reinforcement learning stage based on Group Relative Policy Optimization (GRPO) to enhance role specialization and alignment. Experiments show that \alg outperforms a wide range of existing RL-based and multi-agent methods. In addition, we demonstrate that \alg can generalize to other multi-agent frameworks (e.g., Retrieval and Debugging agents), highlighting its flexibility and scalability. We evaluate \alg on multiple benchmarks of varying difficulty using three base models. Compared to existing baselines, the results show consistent improvements in accuracy, while also achieving higher efficiency in terms of execution time and memory usage, demonstrating the effectiveness and practicality of \alg.
\end{abstract}

\section{Introduction}

With the rapid development of large language models (LLMs) in recent years, LLM-powered code generation methods have shown remarkable capabilities in a wide range of code generation tasks \citep{liu2024empirical, fried2023incoder, koziolek2024llm, li2024starcoder,alomar2024refactor}. LLMs with advanced reasoning abilities, including DeepSeek \citep{guo2025deepseek,guo2024deepseek}, LLaMA \citep{touvron2023llama}, Qwen \citep{hui2024qwen2}, and GPT \citep{singh2025openai}, have achieved notable results on a wide range of code generation benchmarks.
During the generation process, these models typically follow an autoregressive decoding strategy, predicting the next token one step at a time based on the previously generated tokens. This sequential generation mechanism can be viewed as a token-level Markov process \citep{liu2025understanding,wan2025srpo,yao2023tree,dou2025plan}. However, such a generation mechanism may inherently lack global planning. As a result, it often produces outputs that, while locally coherent, are suboptimal solutions from a global perspective.

\begin{figure*}[t]
    \centering
    \subfloat[]{\includegraphics[width=0.245\textwidth]{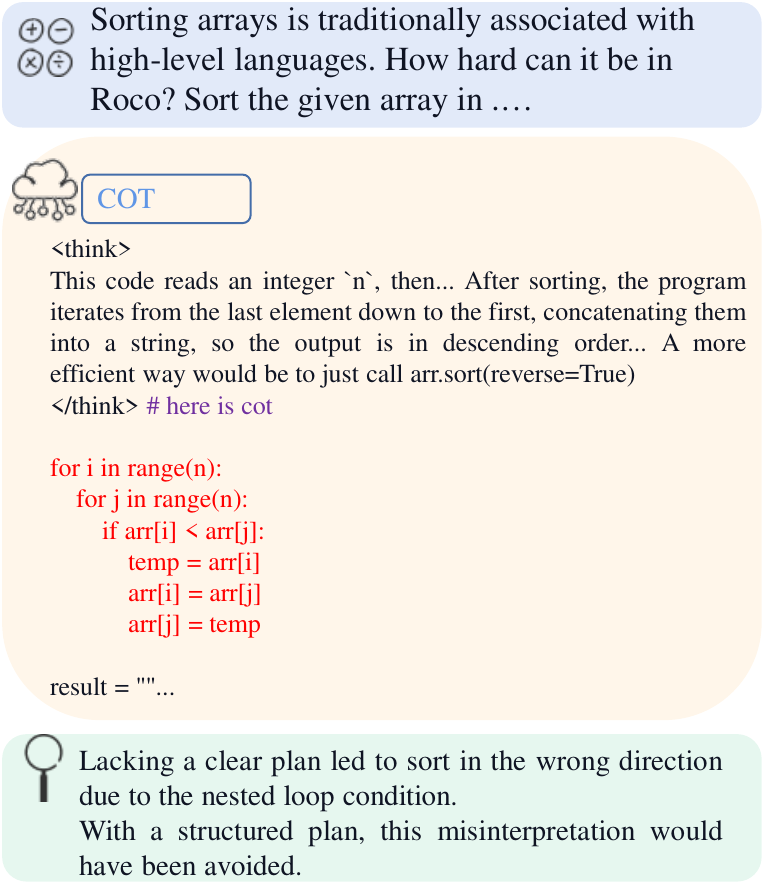}\label{fig:cot-example}}
    \hfill   
    \subfloat[]{\includegraphics[width=0.245\textwidth]{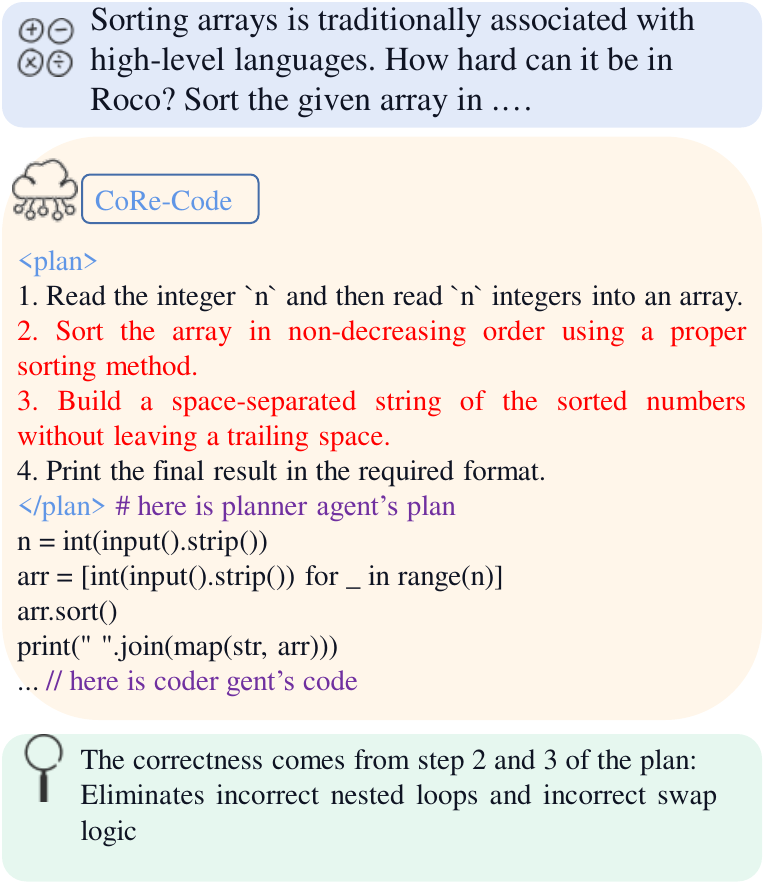}\label{fig:ours-example}}
    \hfill   
    \subfloat[]{\includegraphics[width=0.245\textwidth]{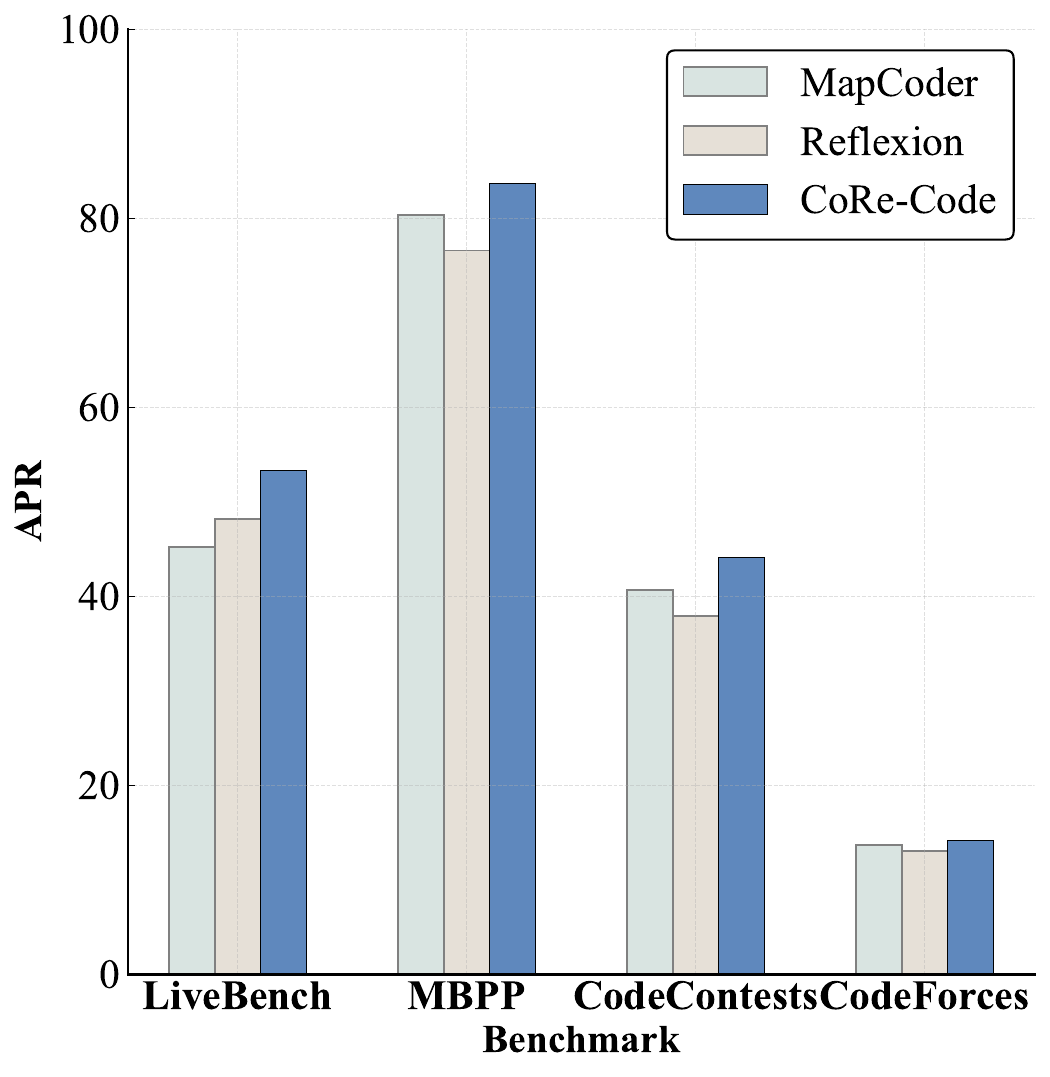}\label{fig:ours-example}}
    \hfill   
    \subfloat[]{\includegraphics[width=0.245\textwidth]{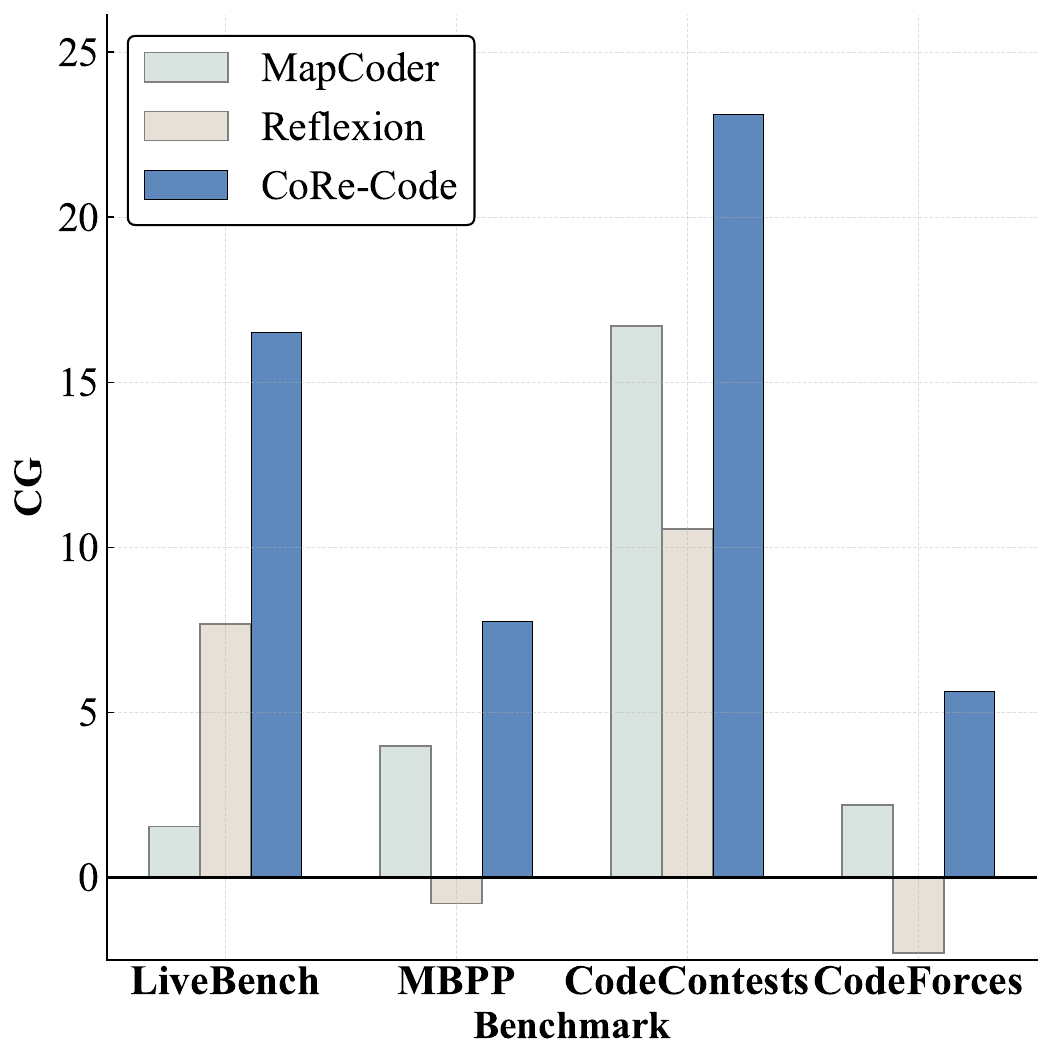}\label{fig:ours-example_CG}}

    \caption{(a) CoT for code generation. (b) \alg for code generation. (c) Comparison with different multi-agent systems for code generation. 
(d) Collaboration Gain values across different models. 
Both (c) and (d) use Qwen2.5-7B-Coder-Instruct as the base model. More examples can be found in Appendix \ref{sec:test_case}.}
    \label{fig:difference}
\end{figure*}

To address this issue and improve the accuracy of code generation, prior studies have proposed the Chain-of-Thought (CoT) method \citep{wei2022chain}, which introduces intermediate reasoning steps or pseudocode before generation to enhance planning and reduce errors. However, CoT struggles with complex, multi-faceted tasks. To overcome this limitation, researchers have further explored multi-agent systems (MAS) based on large language models, where different agents (e.g., requirements engineers, programmers, and testers) collaborate to accomplish end-to-end software development workflows \citep{islam2024mapcoder,lin2025soen}.
However, we observe that existing multi-agent approaches often yield only marginal improvements in code generation performance as shown in Fig \ref{fig:ours-example}. To understand this phenomenon, we analyze the role of collaboration in multi-agent systems. In typical frameworks, the final code is generated by a Coder Agent, while other agents provide intermediate guidance or feedback. We therefore ask: \textit{how much do these auxiliary agents actually help the Coder Agent?} To quantify this effect, we introduce collaboration gain (CG) in Definition \ref{sec:cg}, a metric that measures the effective contribution of auxiliary agents to the Coder Agent’s performance.
As illustrated in Fig \ref{fig:ours-example_CG}, existing multi-agent approaches often fail to produce substantial collaboration gains, and in some cases may even lead to negative gains. This observation indicates that enhancing the collaboration capability among agents remains a critical challenge.

To address these challenges, we propose \textbf{\alg}, a \textbf{Co}llaboration-aware \textbf{Re}inforcement learning framework for multi-agent \textbf{Code} generation, designed to enhance cooperation among agents during code synthesis.
Existing studies on multi-agent reinforcement learning for code generation remain largely underexplored, leaving substantial room for exploring how RL can improve collaboration among specialized coding agents.
We study this problem under the representative Planner--Coder paradigm \citep{lyu2025testing,huang2023agentcoder}, where a Planner agent produces a high-level solution plan and a Coder agent translates it into executable code.
To make planning more concrete and actionable, we introduce \textit{algorithmic thoughts}, a structured representation that decomposes algorithmic reasoning into input--output definition, linear progression, conditional logic, and iteration, inspired by prior work \citep{li2025structured, le2023codechain, chen2022program}.
Nevertheless, directly optimizing the Planner is challenging, as intermediate plan quality is difficult to verify in isolation. Existing alternatives, such as reward model-based supervision \citep{yu2025reward,zhang2024accessing,feng2025prm} and off-policy training from collected planning trajectories \citep{lightman2023let,rafailov2024scaling}, remain unreliable: the former may cause proxy-reward mismatch, where plan-level scores fail to align with final code correctness or executability \citep{yu2025reward,gao2023scaling}, while the latter may suffer from distribution shift between collected plans and current Planner--Coder interaction dynamics.
This motivates a verifiable optimization framework for training the Planner. 
We therefore use downstream code-execution results as verifiable rewards, following Reinforcement Learning with Verifiable Rewards (RLVR), and adopt Group Relative Policy Optimization (GRPO) \citep{shao2024deepseekmath} for policy optimization. 
Based on this design, we propose \textbf{Collaborative GRPO}, a collaboration-aware RL framework that trains the Planner and Coder with role-specific objectives while coordinating their learning through shared execution-based feedback. 
Consequently, the execution results of the Coder serve as indirect yet verifiable feedback for the Planner's outputs. This enables the Planner to learn plans that better support downstream code synthesis, while encouraging the Coder to better align with and execute the generated plans.
To the best of our knowledge, \alg represents one of the first verifier-guided RL-based multi-agent frameworks for code generation, leveraging verifiable execution feedback to improve collaboration among role-specialized agents.
Furthermore, we experimentally show that \alg is not limited to the Planner--Coder paradigm. It can be further extended to optimize other agents, such as the Retrieval Agent and Debugging Agent, demonstrating its generality and flexibility across diverse multi-agent code generation frameworks.

\section{Related work}
Due to space limitations, additional related work  and preliminary knowledge of GRPO are presented in the Appendix \ref{rw:supp}.

\subsection{LLM-based code generation}
The automatic generation of code or completion of snippets from natural language using LLMs has gained much attention \citep{guo2024deepseek,wei2022chain,zhang2023planning,islam2024mapcoder,jiang2024self,izadi2024language,lin2025soen,zhang2025sealign,zhang2024codedpo,rasheeda2026llm,hasanli2026tdd}, improving efficiency and reducing human error \citep{huang2023towards,geng2024large}. Models such as GPT-4o \citep{achiam2023gpt}, ChatGLM \citep{glm2024chatglm}, CODEX \citep{pasquini2010codex}, Qwen \citep{hui2024qwen2}, DeepSeek \citep{guo2025deepseek}, and CodeGen \citep{nijkamp2022codegen} show strong code generation and understanding abilities, achieving SOTA results on MBPP \citep{austin2021program} and HumanEval \citep{chen2021evaluating}. Their success stems from large-scale training \citep{lozhkov2024starcoder} and SFT for better coding abilities \citep{chang2024survey}.

Prompt-based methods further enhance performance. CoT \citep{wei2022chain} generates reasoning steps to guide code; retrieval-based prompting incorporates relevant examples \citep{nashid2023retrieval,kang2023large}; ChatUniTest locates focal methods for test generation \citep{xie2023chatunitest}; prompt composition adds high-level descriptions before code generation \citep{yuan2023no}; and CodeT \citep{chen2022codet} leverages self-generated tests.

Recent training-time methods improve code LLMs through reinforcement learning and curriculum learning. RECRL enhances curriculum RL by modeling requirement difficulty and adaptive sampling \citep{yin2026recrl}, while AgentConductor \citep{wang2026agentconductor} optimizes multi-agent code generation via difficulty-aware topology evolution. SecCoderX \citep{wu2026secure} further applies online RL with vulnerability reward models to improve secure and functional code generation.

\begin{figure*}
    \centering
    \includegraphics[width=0.99\linewidth]{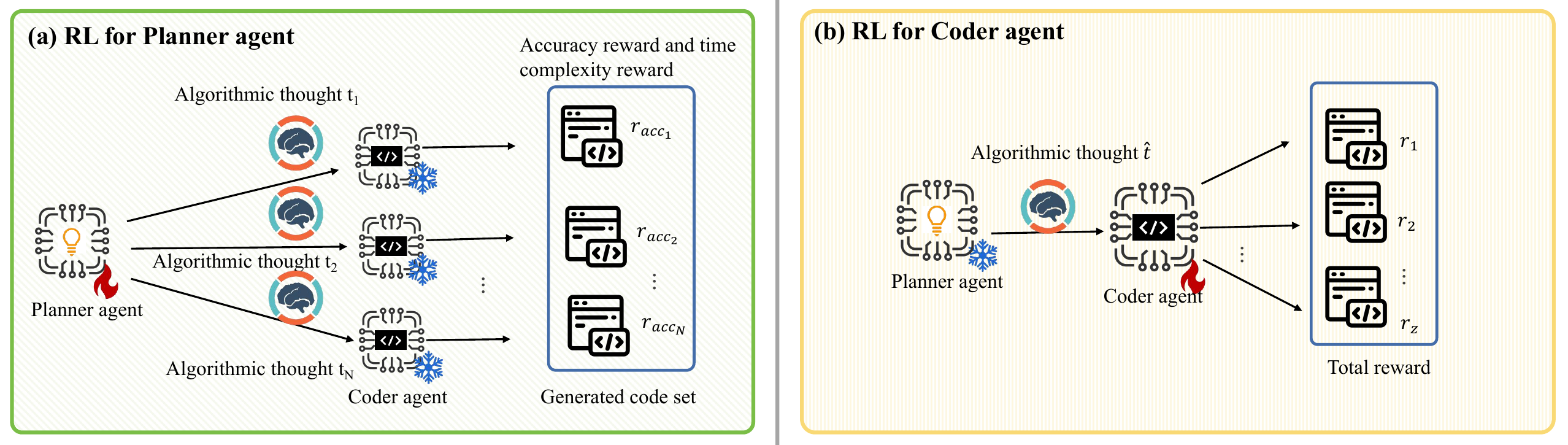}
    \caption{Overview of \alg. (a) The Planner agent is optimized to generate effective algorithmic thoughts, while (b) the Coder agent is optimized to translate the given thought into correct and efficient code.}
    \label{fig:enter-label}
\end{figure*}

\section{Approach}

In this section, we present the methodology of the \alg framework. The framework is built upon a collaboration-aware reinforcement learning approach, termed \textbf{collaborative GRPO}. This two-stage GRPO algorithm jointly optimizes the planner and the coder, encouraging them to collaborate toward a unified objective, thereby improving both the accuracy and efficiency of code generation.
Specifically, the planner is optimized via reinforcement learning to produce more efficient and structured algorithmic plans. The coder, in turn, is rewarded based on how faithfully and correctly it translates the generated plans into executable code, ensuring alignment and coordinated behavior between the two agents. Empirical analysis further demonstrates that this collaborative mechanism enhances inter-agent cooperation and complementarity. The overall process is illustrated in Fig.~\ref{fig:enter-label}.

\begin{definition}[Collaboration Gain]
Let $P_{\text{coder}}(c_i \mid q_i)$ denote the average pass rate
of the coder agent with the initial question $q_i$, and
$P_{\text{coder}}(c_i \mid \theta_{\text{auxiliary}}, q_i)$ denote the pass rate when
the coder is conditioned on both other auxiliary agents $\theta_{\text{auxiliary}}$ and $q_i$.
The collaboration gain (CG) is defined as
\begin{equation}
CG = 1 - \frac{P_{\text{coder}}(c_i \mid q_i)}
           {P_{\text{coder}}(c_i \mid \theta_{\text{auxiliary}}, q_i)} .
\end{equation}
\label{sec:cg}
\end{definition}

Since pass rates lie in $[0,1]$, the collaboration gain satisfies
$CG \le 1$. A larger $CG$ indicates that the other auxiliary agent
$\theta_{\text{auxiliary}}$ provides stronger support for the coder agent and leads to
greater performance improvement, whereas a smaller $CG$ suggests
limited or negligible collaborative benefit. When CG is less than 0, it indicates that the auxiliary agent has a negative effect.

\subsection{Better planning structure via Algorithmic Thought}
\label{sec:Algorithmic_Thought}

\begin{wrapfigure}{r}{0.55\textwidth}
    \centering
    \vspace{-10pt}
    \includegraphics[width=0.46\textwidth]{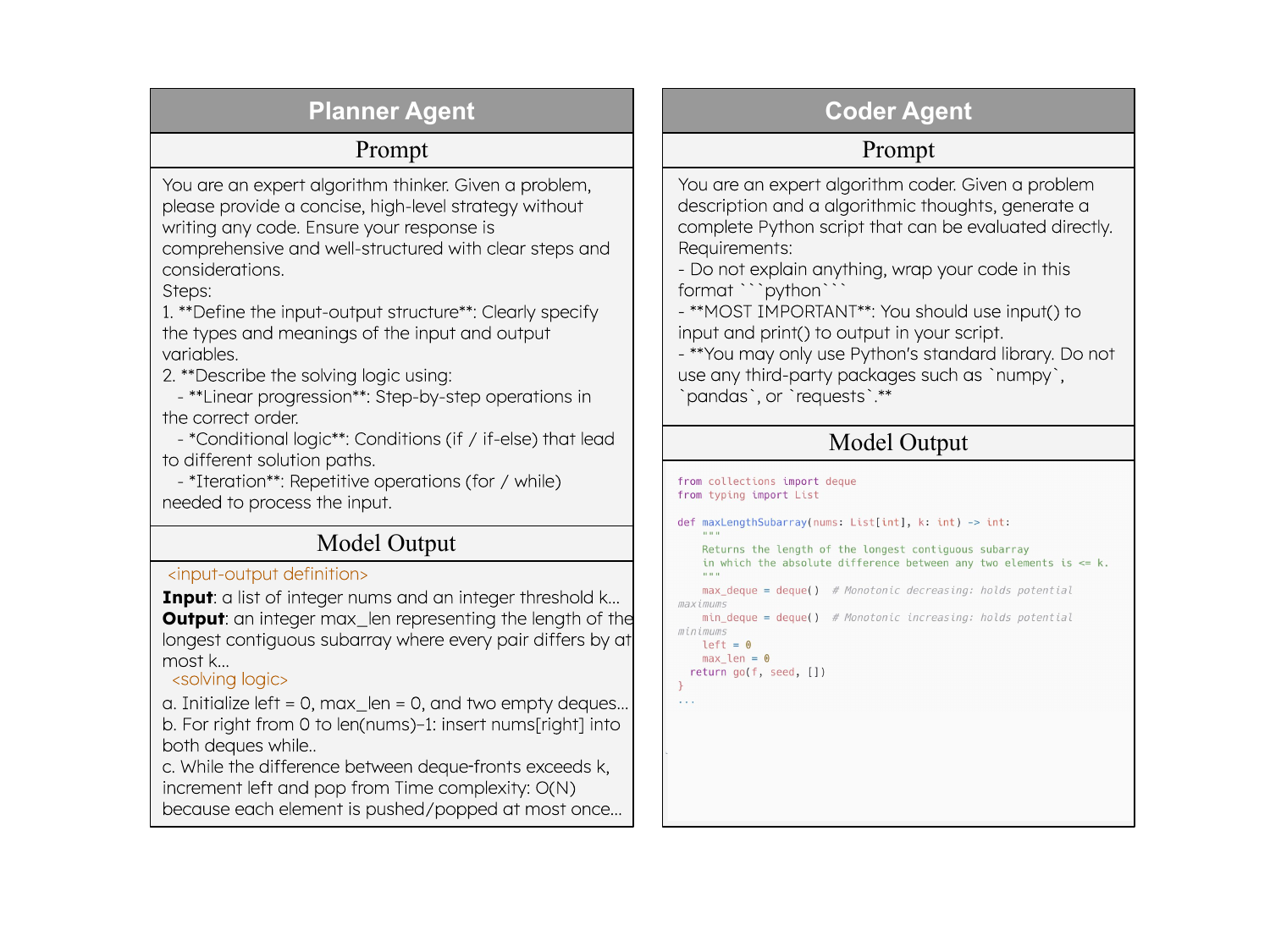}
    \caption{Algorithmic Thought example. The Planner Agent prompt is shown on the left, and the Coder Agent prompt is shown on the right.}
    \label{fig:example}
    \vspace{-10pt}
\end{wrapfigure}
To encourage the Planner Agent to generate more effective and code-oriented plan structures, we propose \textbf{Algorithmic Thought}, a structured planning representation tailored for program synthesis. Inspired by \citep{li2025structured,le2023codechain,chen2022program}, Algorithmic Thought organizes the planner's output into four core components: \textbf{input-output definition}, \textbf{linear progression}, \textbf{conditional logic}, and \textbf{iteration}. The input-output definition clarifies the functional objective by specifying the available inputs and the expected outputs. Linear progression outlines the sequential computational steps that transform inputs into outputs. Conditional logic specifies branch decisions under different cases, while iteration describes repeated computation over data structures or ranges. This structured formulation makes the planner's reasoning more aligned with programming semantics, improves the interpretability of intermediate plans, and provides the Coder Agent with clearer and more executable guidance for generating accurate and coherent code. Fig.~\ref{fig:example} illustrates an example of Algorithmic Thought together with the prompt produced by the Planner Agent.

\subsection{Collaborative GRPO Reinforcement Learning Framework}

We propose \textbf{Collaborative GRPO}, a verifiable-reward extension of the Group Relative Policy Optimization (GRPO) reinforcement learning algorithm \citep{guo2025deepseek}, to jointly optimize both agents using downstream code-execution feedback. By grounding policy optimization in verifiable execution results rather than unverifiable proxy rewards, this collaborative learning mechanism enhances the specialization and coordination of the Planner and Coder, further increasing their collaboration gain and enabling more efficient code generation.%


\subsubsection{RL for Planner Agent Specialization} 
\label{sec:planner_grpo}

The specialization of planner agents falls under the collaborative RL paradigm. For the optimization process of the planner agent, a key challenge is that algorithmic thought's quality cannot be directly measured or verified. Therefore, we leverage the cooperative interaction between the planner and coder agents: the planner generates algorithmic thoughts, while the coder translates them into executable code. The resulting code execution outcomes provide verifiable signals, which are then used to optimize the planner parameters \( \theta_{\text{planner}} \).


Given a question $q$, the planner agent first generate $N$ distinct algorithmic thoughts $\{t_i\}_{i=1}^N$, where $t_i =\pi_{\theta_{\text{planner}}}(q)$. To evaluate the quality of each algorithmic thought, we adopt an indirect metric: the accuracy of code generated by the Coder Agent under the policy $\pi_{\theta_{\text{coder}}}$. Specifically, for a given thought $t_i$, we pair it with the question $q$ and input them into the Coder Agent to produce $M$ candidate code snippets $\{c_{i,j}\}_{j=1}^M$, where each code snippet is generated as $c_{i,j} = \pi_{\theta_{\text{coder}}}(t_i, q)$. The more of the question’s provided test cases that the set \( \{c_{i,j}\}_{j=1}^M \) successfully passes, the higher the quality we ascribe to the corresponding algorithmic thought \( t_i \).
Therefore, the accuracy reward $r_{\text{acc}_i}$ for algorithmic thought $t_i$ can be expressed as:
{\small
\begin{equation}
r_{\text{acc}_i} = \frac{1}{M} \sum_{j=1}^{M} \sigma\!\left(p_{i,j}\right)\quad \text{where} \quad p_{i,j} = \frac{1}{|\mathcal{T}|}
\sum_{k=1}^{|\mathcal{T}|}
\mathbb{I}\left[
\mathrm{Exec}(c_{i,j}, x_k) = y_k
\right]. 
\label{acc_reward}
\end{equation}}
$ \sigma(\cdot)$ is scaled sigmoid function, \( p_{i,j} \) denotes the proportion of test cases passed by the code snippet \( c_{i,j} \), 
\(\mathrm{Exec}(c_{i,j}, x_k)\) is the execution output of \(c_{i,j}\) on input \(x_k\), 
and \(y_k\) is the ground-truth output of the \(k\)-th test case. The accuracy reward \( r_{\text{acc},i} \) evaluates multiple candidate code snippets derived from the same algorithmic thought \( t_i \). Instead of directly averaging the raw pass rates \( p_{i,j} \), we use a \text{sigmod}-based weighting scheme to assign larger weights to higher-quality snippets that pass more test cases. As a result, algorithmic thoughts that produce stronger code receive higher rewards, while those leading to weaker solutions are down-weighted. This provides a more informative and discriminative reward signal for assessing the quality of algorithmic reasoning.
Meanwhile, to approximately assess the algorithmic complexity induced by the plan $t_i$, we introduce a time complexity reward $r_{\text{time}_i}$, which is estimated based on the complexity of the generated code snippet set $\{c_{i,j}\}_{j=1}^{M}$. Specifically, following \citep{goldsmith2007measuring}, we compute the time complexity of each generated code snippet $c_{i,j}$, which is denoted as $T_{i,j}$. The time complexity reward for the algorithmic thought $t_i$ is defined as:
$
r_{\text{time}_i} =
\frac{1}{M}
\sum_{j=1}^{M}
\sigma(-{T_{i,j}}).
$
Intuitively, this formulation assigns higher rewards to algorithmic thoughts that lead to code with lower time complexity, thereby encouraging the planner to generate more efficient solutions. The time complexity prediction algorithm of $r_{\text{time}_i}$, presented in Algorithm \ref{algo:time_complexity_prediction}, is shown in Appendix \ref{sec:complex}.
After obtaining the accuracy rewards \( \{ r_{\text{acc},i} \}_{i=1}^n \) and time complexity rewards \( \{ r_{\text{time},i} \}_{i=1}^n \) for all candidate thoughts, the total reward for each thought is computed as
$
r_{\text{total},i} = r_{\text{time},i} + r_{\text{acc},i}.
$
Based on the total reward set $\{r_{\text{total},i}\}_{i=1}^n$,
we compute the advantage function using Eq.~\ref{advantages} and update the planner parameters \( \pi_{\theta_{\text{planner}}} \) according to Eq.~\ref{grpo}.

During this RL stage, the Planner and Coder Agents collaborate to construct verifiable reward signals that are used solely to update the Planner’s parameters. Rather than relying on subjective or hard-to-quantify assessments of plan quality, we convert the quality of a generated plan into measurable and verifiable downstream feedback through the Coder Agent. In this way, the Planner is optimized to generate higher-quality algorithmic thoughts that better support the Coder in producing accurate and efficient code, while the Coder remains frozen throughout this stage.
In the subsequent RL stage for Coder specialization, its parameters are updated, as detailed in Section~\ref{sec:codeRL}.

\subsubsection{RL for Coder Agent Specialization}
\label{sec:codeRL}

After RL for planner specialization, the specialized planner agent becomes capable of generating higher-quality algorithmic thoughts. In the subsequent stage, the focus of RL shifts to coder agent specialization, aiming to enhance the coder agent's ability to effectively follow these algorithmic thoughts and produce accurate, efficient code.

For a given question $q$, we first use the reinforcement-trained planner agent to generate an algorithmic thought $\hat{t}$, where $\hat{t} = \pi_{\theta_{\text{planner}}}(q)$. Guided by the algorithmic thought $\hat{t}$, the coder agent generates a set of code snippets ${c_i}_{i=1}^{z}$, consisting of $z$ code snippets. This process can be formulated as $\{c_i\}_{i=1}^{z} = \pi_{\theta_{\text{coder}}}(q, \hat{t})$. 
In a similar manner, based on the test case pass rate defined in Eq.~\ref{acc_reward}, we calculate the accuracy reward for each code snippet, resulting in the reward set $\{r_{\text{acc}_i}\}_{i=1}^z$.

The objective of the coder agent is not only to generate accurate code, but also to produce an efficient implementation. For a set of code snippets $\{c_i\}_{i=1}^z$, we assign a memory efficiency reward if at least one code snippet passes all test cases; if no snippet passes, the efficiency reward is set to $0$. We employ package \texttt{psutil} to monitor both storage space and memory usage.
Among all snippets that pass every test case, the one with the smallest memory consumption is defined to have the target memory usage, denoted as $\mathcal{O}_{\text{target}}$.
The value of each space efficiency reward  $r_{\text{space}_i}$ of code snippet $c_i$ is determined as:



{\small\begin{equation}
r_{\text{space}_i} =
\begin{cases}
\exp\!\left(-\bigl|\mathcal{O}(c_i)-\mathcal{O}_{\text{target}}\bigr|\right), & \text{if } c_i \text{ passes all test cases}, \\[8pt]
0, & \text{otherwise}.
\end{cases}
\end{equation}}

where $\mathcal{O}(c_i)$ represents the space complexity of the code snippet $c_i$.
The space efficiency reward is only applicable when the generated code snippet passes all test cases. 
Its purpose is to ensure that while rewarding the accuracy of code generation, the coder agent is also gradually encouraged to focus on producing implementations whose space complexity approaches $\mathcal{O}_{\text{target}}$.
We introduce the space efficiency reward $r_{\text{space}_i}$ during the coder agent's RL stage to constrain the concrete implementation. As the algorithmic thought $t_i$ specifies only a high-level strategy, implementations under the same $t_i$ may exhibit different space complexities. The reward therefore encourages solutions approaching the target complexity $\mathcal{O}_{\text{target}}$.
Finally, our total reward $r_i$ for a code snippet is defined as:
\begin{equation}
    r_i = r_{\text{acc}_i} + \lambda r_{\text{space}_i},
\end{equation}
where $\lambda$ is a hyperparameter.
After obtaining the total reward \( r_i \), we compute the advantage function based on Eq.~\ref{advantages}, and update the parameters of the coder agent according to Eq.~\ref{grpo}. Through the RL for Coder Agent stage, the Coder Agent learns to better align with the Planner’s algorithmic thoughts and faithfully execute the provided instructions. As a result, it can produce code with improved correctness and computational efficiency.

\subsection{Empirical analysis of collaboration gain in training process}

\begin{figure*}[h]
    \centering
    \subfloat[Qwen2.5-7B-Coder-Instruct]{\includegraphics[ width=0.33\textwidth]{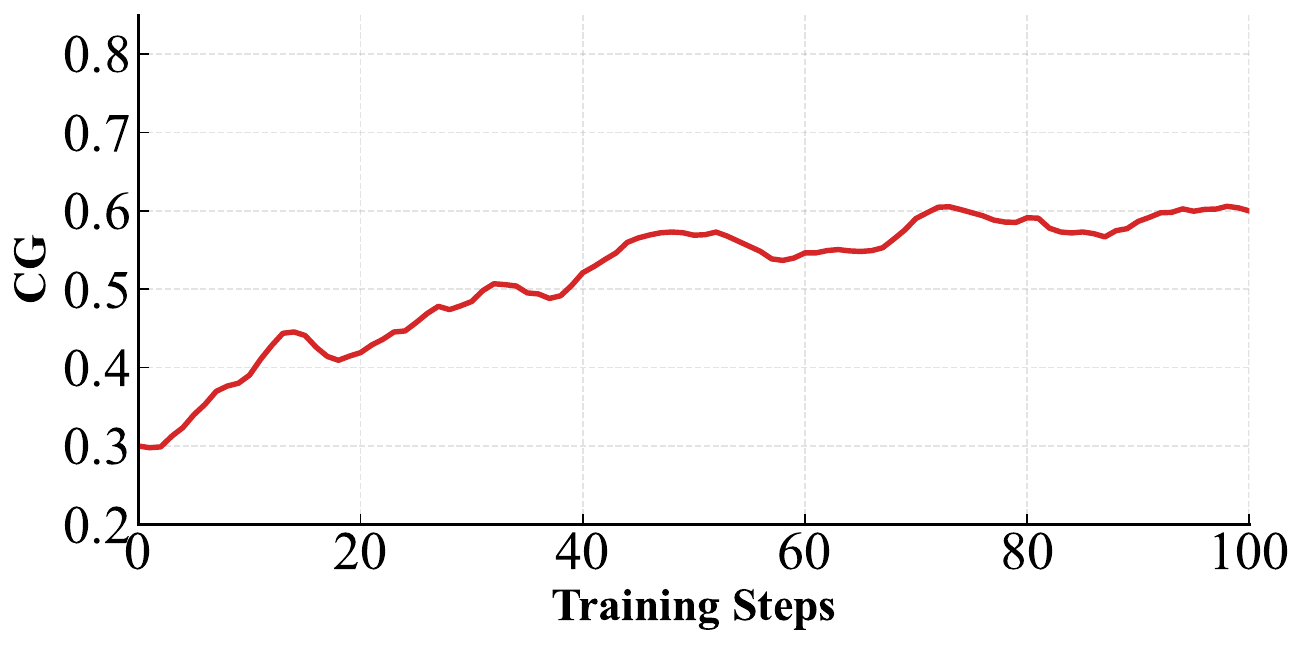}\label{CG_QWEN2.5_7b}}
    \hfill   
    \subfloat[Qwen2.5-14B-Coder-Instruct]{\includegraphics[width=0.33\textwidth]{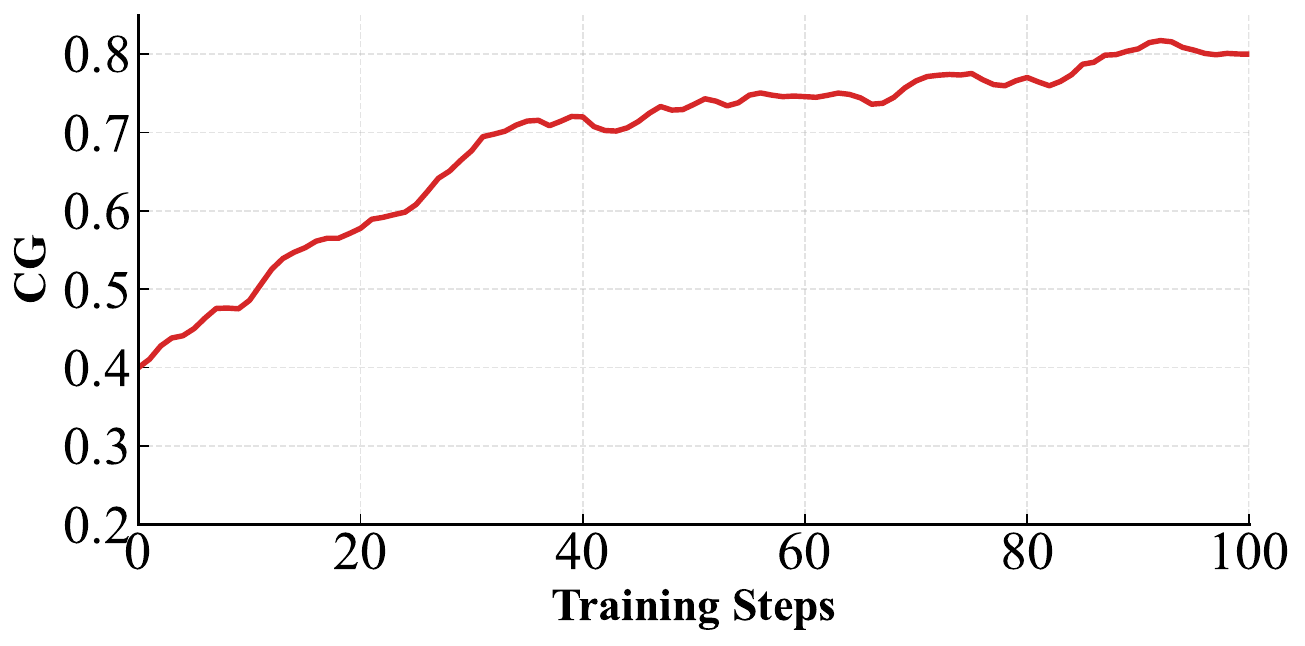}\label{CG_QWEN2.5_14b}}
    \hfill   
    \subfloat[Qwen3-4B]{\includegraphics[width=0.33\textwidth]{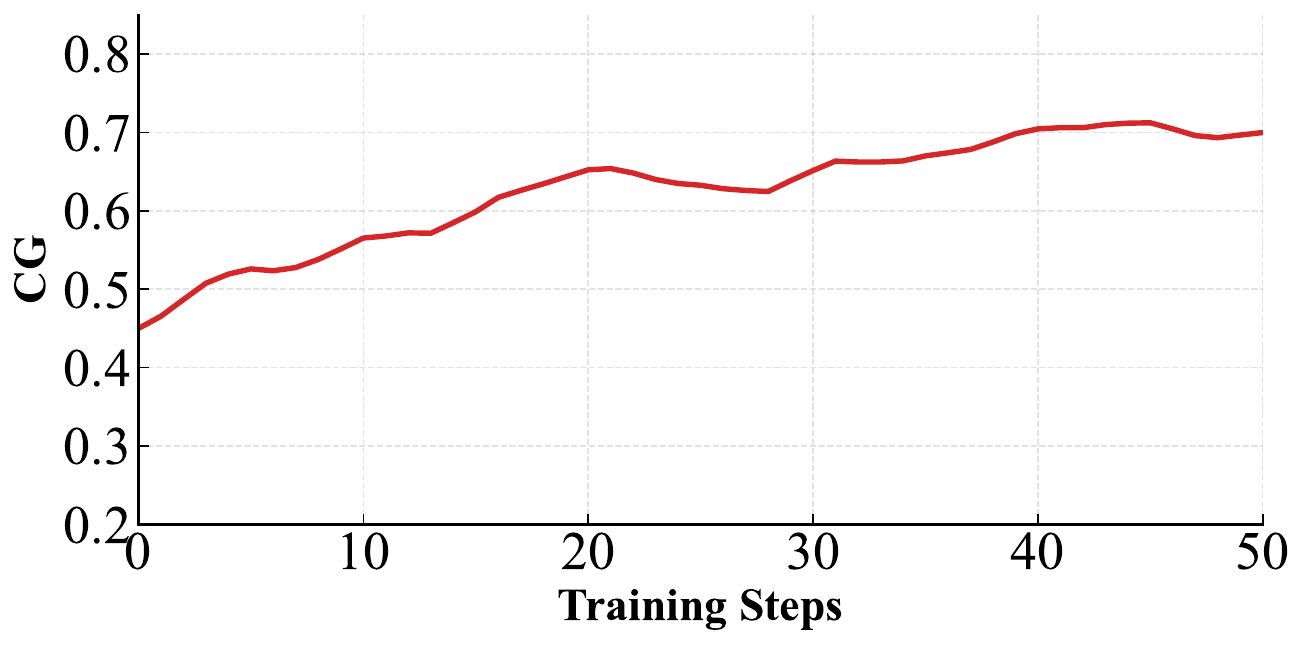}}\label{CG_QWEN3_4b}

    \caption{RL training dynamics of Collaboration Gain for Planner agent across different models.}
    \label{fig:difference}
\end{figure*}

Figure~\ref{fig:difference} illustrates the training dynamics of Collaboration Gain (CG) when applying \alg to optimize the Planner agent. We observe a clear upward trend of CG across all three base models as training progresses, including Qwen2.5-7B-Coder-Instruct, Qwen2.5-14B-Coder-Instruct, and Qwen3-4B. This result indicates that \alg progressively improves the coordination between the Planner and the other agents during RL training. In particular, the increasing CG suggests that the Planner learns to produce more effective high-level plans that better support downstream code generation, thereby strengthening inter-agent collaboration. Although the growth trajectories differ slightly across models, the overall trend remains consistent, demonstrating that \alg can steadily enhance collaborative effectiveness in the multi-agent system.

\section{Experiment}
\label{sec:exp}

\subsection{Experiment setup}
The detailed RL training parameters can be found in Appendix~\ref{sec:details}.

\subsubsection{Base model and benchmark}

Since \alg requires additional training process of LLMs, this paper focuses on open-source models.
In our experiments, we choose Qwen2.5-7B-Coder-Instruct, Qwen2.5-14B-Coder-Instruct \citep{hui2024qwen2} and Qwen 3-4B \citep{yang2025qwen3} as the base models for \alg.
We evaluate them on four different benchmarks: LiveCode \citep{white2024livebench}, MBPP \citep{austin2021program}, CodeContests \citep{li2022competition} and CodeForces \citep{quan2025codeelo}. LiveCode and MBPP are classified as basic function-level programming tasks, which focus on evaluating the fundamental programming skills of a model, while CodeContests and CodeForces are classified as complex programming tasks of competition level, which focus on evaluating the advanced algorithm design and complexity management capabilities of a model.

\subsubsection{Metrics}


For code generation accuracy, we adopt Pass@1, Pass@5, and Average Pass Rate (APR) \citep{li2025structured,wang2025co,zhang2024unseen}. Pass@1 checks correctness on the first attempt, Pass@5 allows up to five attempts, and APR measures the average proportion of passed test cases.
For efficiency, we record average runtime and memory usage (MU, in char).
For maintainability and correctness, we compute average cyclomatic complexity (CC) and failure rate (FR).
Finally, we measure inference time during code generation.


\setlength{\tabcolsep}{2.5pt}   
\renewcommand{\arraystretch}{0.95}
\begin{table*}[]
\caption{Performance on various benchmarks using Pass@1 ($\uparrow$), Pass@5 ($\uparrow$), and APR ($\uparrow$) across different base models. Q2.5-7B-I, Q2.5-7B-CI, Q3-4B, Q2.5-14B-CI, and DSCV2-16B denote 
Qwen-7B-Instruct, Qwen-Coder-7B-Instruct, Qwen3-4B, 
Qwen-Coder-14B-Instruct, and DeepSeek-Coder-V2-16B, respectively. \textbf{Bold} indicates the best-performing result. }
\centering
\scriptsize
\begin{tabular}{c|ccccccccccccc}
\midrule[1.2pt]
\multirow{2}{*}{Based model} & \multirow{2}{*}{Method} & \multicolumn{3}{c}{LiveBench}                            & \multicolumn{3}{c}{MBPP}                                 & \multicolumn{3}{c}{CodeContests}                         & \multicolumn{3}{c}{CodeForces}                           \\ \cline{3-14} 
                             &                         & Pass@1        & Pass@5        & APR                      & Pass@1        & Pass@5        & APR                      & Pass@1        & Pass@5        & APR                      & Pass@1        & Pass@5        & APR                      \\ \hline
Q2.5-7B-I                       & -                       & 28.3          & 35.2          & 41.9                     & 63.6          & 68.7          & 71.4                     & 19.4          & 21.9          & 27.2                     & 5.6           & 5.9           & 7.7                      \\
Q2.5-7B-CI                      & -                       & 32.8          & 37.5          & 44.5                     & 67.7          & 74.4          & 77.2                     & 21.3          & 24.4          & 33.9                     & 6.2           & 8.3           & 13.4                     \\
Q3-4B                        & -                       & 36.9          & 41.4          & 42.4                     & 72.7          & 75.4          & 77.8                     & 26.3          & 31.5          & 27.7                     & 7.8           & 13.4          & 10.7                     \\
Q2.5-14B-CI                     & -                       & 40.6          & 47.7          & 53.1                     & 78.3          & 81.9          & 85.0                     & 29.1          & 33.1          & 41.4                     & 9.2           & 13.5          & 17.5                     \\
DSCV2-16B                    & -                       & 23.4          & 26.4          & 36.5                     & 71.9          & 77.8          & 77.9                     & 21.3          & 28.5          & 34.4                     & 7.5           & 9.6           & 16.7                     \\ \hline
\multirow{5}{*}{Q3-4B}       & GRPO                    & 39.2          & 42.7          & 49.7                     & 78.5          & 80.3          & 81.8                     & 28.7          & 30.3          & 44.2                     & 13.5          & 15.5          & 15.0                     \\
                             & Focused-DPO             & 40.5          & 44.2          & 50.5                     & 77.2          & 79.4          & 81.2                     & 25.4          & 30.3          & 29.8                     & 14.4          & \textbf{17.1} & 18.7                     \\
                             & CURE                    & 38.5          & 40.7          & 50.4                     & 75.7          & 79.3          & 79.7                     & 24.2          & 28.7          & 33.9                     & 13.2          & 15.3          & 16.4                     \\
                             & CodeRL+                 & 36.9          & 44.2          & 47.4                     & 74.5          & 77.4          & 80.2                     & 28.3          & 33.2          & 37.4                     & 14.2          & 14.9          & 16.7                     \\
                             & \cellcolor{greyL}\alg     & \cellcolor{greyL}\textbf{42.4} & \cellcolor{greyL}\textbf{44.9} & \cellcolor{greyL}\textbf{54.3}            & \cellcolor{greyL}\textbf{80.5} & \cellcolor{greyL}\textbf{82.4} & \cellcolor{greyL}\textbf{82.2}            & \cellcolor{greyL}\textbf{31.6} & \cellcolor{greyL}\textbf{34.9} & \cellcolor{greyL}\textbf{42.7}            & \cellcolor{greyL}\textbf{15.9} & \cellcolor{greyL}17.0          & \cellcolor{greyL}\textbf{19.4}            \\ \hline
\multirow{5}{*}{Q2.5-7B-CI}  & GRPO                    & 34.5          & 41.2          & 51.3                     & 73.4          & 75.7          & 80.5                     & 24.5          & 28.5          & 43.3                     & 9.1           & 9.7           & 13.3                     \\
                             & Focused-DPO             & 33.8          & 40.3          & 46.4                     & 69.2          & 73.5          & 79.7                     & 22.7          & 27.2          & 36.7                     & 7.7           & 9.5           & 12.9                     \\
                             & CURE                    & 37.1          & 39.3          & 48.7                     & 70.2          & 74.5          & 78.7                     & 25.9          & 29.4          & 39.2                     & 8.2           & 9.4           & 11.9                     \\
                             & CodeRL+                 & 35.3          & 39.5          & \multicolumn{1}{l}{50.7} & 66.9          & 72.7          & \multicolumn{1}{l}{80.8} & \textbf{28.5} & 30.3          & \multicolumn{1}{l}{40.7} & 7.4           & 10.7          & \multicolumn{1}{l}{12.2} \\
                             & \cellcolor{greyL}\alg     & \cellcolor{greyL}\textbf{37.7} & \cellcolor{greyL}\textbf{42.2} & \cellcolor{greyL}\textbf{53.3}            & \cellcolor{greyL}\textbf{73.9} & \cellcolor{greyL}\textbf{77.2} & \cellcolor{greyL}\textbf{83.7}            & \cellcolor{greyL}27.4          & \cellcolor{greyL}\textbf{32.2} & \cellcolor{greyL}\textbf{44.1}            & \cellcolor{greyL}\textbf{11.7} & \cellcolor{greyL}\textbf{13.4} & \cellcolor{greyL}\textbf{14.2}            \\ \hline
\multirow{5}{*}{Q2.5-14B-CI} & GRPO                    & 45.5          & 50.7          & 55.4                     & 80.7          & 83.2          & 85.2                     & 33.7          & 35.9          & 42.3                     & 12.9          & 14.3          & 17.6                     \\
                             & Focused-DPO             & 41.9          & 46.2          & 55.7                     & 78.4          & 79.3          & 79.2                     & 31.7          & 34.4          & \textbf{47.2}            & 9.7           & 12.3          & 19.4                     \\
                             & CURE                    & 47.5          & 49.3          & 56.7                     & 78.5          & 80.5          & 81.9                     & 32.1          & 36.2          & 43.6                     & 12.1          & 14.1          & 18.8                     \\
                             & CodeRL+                 & 44.7          & 52.4          & 57.9                     & 80.4          & 82.7          & 85.5                     & 29.2          & 33.5          & 47.7                     & 10.4          & 12.2          & 21.7                     \\
                             & \cellcolor{greyL}\alg     & \cellcolor{greyL}\textbf{48.4} & \cellcolor{greyL}\textbf{53.5} & \cellcolor{greyL}\textbf{60.4}            & \cellcolor{greyL}\textbf{82.2} & \cellcolor{greyL}\textbf{84.7} & \cellcolor{greyL}\textbf{86.5}            & \cellcolor{greyL}\textbf{34.6} & \cellcolor{greyL}\textbf{36.4}          & \cellcolor{greyL}44.5                     & \cellcolor{greyL}\textbf{13.3} & \cellcolor{greyL}\textbf{15.9} & \cellcolor{greyL}\textbf{22.4}            \\ \midrule[1.2pt]
\end{tabular}
\label{main_results}
\end{table*}

\subsubsection{Baselines}

For a fair comparison, we select four representative RL-based methods as baselines, including GRPO \citep{shao2024deepseekmath}, Focused-DPO \citep{zhang2025focused}, CURE \citep{wang2025co}, and CodeRL+ \citep{jiang2025coderl+}. All RL-based methods, including \alg, are trained on the same training data to ensure a consistent experimental setting. we adopt the reward function proposed in \citep{robeyns2025improving} to optimize the base model via RL. In addition, we compare \alg with three multi-agent code generation methods, namely SCoT \citep{li2025structured}, Reflexion \citep{shinn2023reflexion}, and MapCoder \citep{islam2024mapcoder}, to further evaluate its effectiveness against existing agent-based frameworks.

\subsection{Experimental Results}

Due to space limitations, we placed the evaluation of \alg regarding efficiency and maintainability, error rate, and inference time consumption in Appendix \ref{sec:extra_exp}.

\myparatight{Main results}
Table~\ref{main_results} shows that \alg consistently improves code generation performance across different base models and benchmarks. Compared with representative RL-based baselines, \alg achieves the best overall results on most metrics, especially on Pass@1 and APR, indicating stronger single-sample correctness and more reliable ranking quality. The gains are observed not only on relatively standard benchmarks such as MBPP and LiveBench, but also on more challenging competitive programming benchmarks including CodeContests and CodeForces. This suggests that collaboration-aware optimization effectively enhances the interaction between agents and improves both functional correctness and problem-solving robustness across model scales.

\myparatight{Ablation analysis}
Table~\ref{Tab:ablation} presents the ablation results of different components in \alg. Removing all collaboration-aware RL components leads to the weakest performance across most benchmarks, showing that simple planner--coder prompting is insufficient to fully exploit multi-agent collaboration. Introducing either Planner RL or Coder RL improves the results, indicating that both agents contribute positively to the final code generation quality. Moreover, \alg$_{\text{w/ all}}$ achieves the best overall performance on most Pass@1, Pass@5, and APR metrics, especially on MBPP, CodeContests, and CodeForces. These results demonstrate that jointly optimizing the Planner and Coder agents provides complementary benefits and leads to more effective problem solving.

\myparatight{Comparison with different multi-agent systems}
Table~\ref{com_multi_agent} compares \alg with several representative multi-agent code generation methods, including SCoT, Reflexion, and MapCoder. As shown, \alg achieves the best performance across all four benchmarks and consistently outperforms competing methods on Pass@1, Pass@5, and APR. The improvements are especially clear on more challenging benchmarks such as CodeContests and CodeForces, demonstrating that \alg enables more effective collaboration between agents and leads to stronger code generation quality and robustness. These results verify the advantage of our collaboration-aware optimization over existing multi-agent systems.


\begin{table*}[]
\scriptsize
\centering
\setlength{\tabcolsep}{3.7pt}
\caption{Ablation analysis for each component of \alg, where higher Pass@1 (\(\uparrow\)), Pass@5 (\(\uparrow\)), and APR (\(\uparrow\)) indicate better performance. The base model considered is Qwen2.5-7B-Coder-Instruct.
\textbf{Bold} indicates the best performance for clarity.}
\begin{tabular}{ccccccccccccc}
\midrule[1.2pt]
\multirow{2}{*}{Method}                       & \multicolumn{3}{c}{LiveBench}                 & \multicolumn{3}{c}{MBPP}                      & \multicolumn{3}{c}{CodeContests}              & \multicolumn{3}{c}{CodeForces}                \\ \cline{2-13} 
                                              & Pass@1        & Pass@5        & APR           & Pass@1        & Pass@5        & APR           & Pass@1        & Pass@5        & APR           & Pass@1        & Pass@5        & APR           \\ \hline
\alg$_{\text{w/o all}}$        & 33.9          & 39.3          & 47.9          & 68.5          & 72.8          & 77.8          & 22.6          & 29.7          & 38.9          & 8.9           & 9.6           & 13.3          \\
\alg$_{\text{w/o Planner RL}}$ & 35.3          & 40.7          & \textbf{54.2} & 71.3          & 75.4          & 79.5          & 25.4          & 31.5          & 39.4          & 10.2          & 12.4          & 13.7          \\
\alg$_{\text{w/o Coder RL}}$   & 36.7          & 41.4          & 51.7          & 72.9          & 73.7          & 80.9          & 26.7          & 30.2          & 41.2          & 9.7           & 12.0          & 13.3          \\
\cellcolor{greyL}\alg$_{\text{w/ all}}$         & \cellcolor{greyL}\textbf{37.7} & \cellcolor{greyL}\textbf{42.2} & \cellcolor{greyL}53.3          & \cellcolor{greyL}\textbf{73.9} & \cellcolor{greyL}\textbf{77.2} & \cellcolor{greyL}\textbf{83.7} & \cellcolor{greyL}\textbf{27.4} & \cellcolor{greyL}\textbf{32.2} & \cellcolor{greyL}\textbf{44.1} &\cellcolor{greyL} \cellcolor{greyL}\textbf{11.7} & \cellcolor{greyL}\textbf{13.4} & \cellcolor{greyL}\textbf{14.2} \\ \midrule[1.2pt]
\end{tabular}
\label{Tab:ablation}
\end{table*}

\begin{table*}[]
\centering
\small
\caption{Comparison of different multi-agent methods on various benchmarks using Pass@1 ($\uparrow$), Pass@5 ($\uparrow$), and APR ($\uparrow$), where Qwen2.5-7B-Coder-Instruct is used as the base model. \textbf{Bold} indicates the best-performing result.}
\begin{tabular}{ccccccccccccc}
\midrule[1.2pt]
\multirow{2}{*}{Method} & \multicolumn{3}{c}{LiveBench}                 & \multicolumn{3}{c}{MBPP}                      & \multicolumn{3}{c}{CodeContests}              & \multicolumn{3}{c}{CodeForces}                \\ \cline{2-13} 
                        & Pass@1        & Pass@5        & APR           & Pass@1        & Pass@5        & APR           & Pass@1        & Pass@5        & APR           & Pass@1        & Pass@5        & APR           \\ \hline
SCoT                    & 35.3          & 36.5          & 46.9          & 70.2          & 75.5          & 78.4          & 24.2          & 28.4          & 36.5          & 7.7           & 9.7           & 12.7          \\
Reflexion               & 34.7          & 37.2          & 48.2          & 70.9          & 74.7          & 76.6          & 25.7          & 30.3          & 37.9          & 8.2           & 9.6           & 13.1          \\
MapCoder                & 32.7          & 35.9          & 45.2          & 72.3          & 75.7          & 80.4          & 22.8          & 29.5          & 40.7          & 7.4           & 8.4           & 13.7          \\
\cellcolor{greyL}\alg     & \cellcolor{greyL}\textbf{37.7} & \cellcolor{greyL}\textbf{42.2} & \cellcolor{greyL}\textbf{53.3} & \cellcolor{greyL}\textbf{73.9} & \cellcolor{greyL}\textbf{77.2} & \cellcolor{greyL}\textbf{83.7} & \cellcolor{greyL}\textbf{27.4} & \cellcolor{greyL}\textbf{32.2} & \cellcolor{greyL}\textbf{44.1} & \cellcolor{greyL}\textbf{11.7} & \cellcolor{greyL}\textbf{13.4} & \cellcolor{greyL}\textbf{14.2} \\ \midrule[1.2pt]
\end{tabular}
\label{com_multi_agent}
\end{table*}


\subsection{Extension to other multi-agent framework}
\label{sec:exten}

To evaluate the flexibility of \alg beyond the Planner--Coder paradigm, we further extend it to MapCoder \citep{islam2024mapcoder}, a representative multi-agent code generation framework with Retrieval, Planning, Coding, and Debugging agents. The Planning and Coding agents are kept fixed throughout the experiment. We then train the Retrieval Agent and the Debugging Agent separately with our collaboration-aware RL, where only the target agent is updated and all remaining agents are frozen. For the Retrieval Agent, \alg encourages the generation of more useful self-retrieved exemplars, including relevant problems, plans, code solutions, and algorithmic hints, using downstream code execution performance as a verifiable reward. For the Debugging Agent, \alg optimizes code revision based on the problem, sample I/O feedback, execution logs, and the fixed plan, with rewards assigned according to whether the revised code passes the provided test cases.

\begin{table}[]
\caption{Extension of \alg to MapCoder by independently reinforcing the Retrieval and Debugging agents. We use Qwen2.5-7B-Coder-Instruct as the base model for both agents, while keeping the Planning and Coding agents fixed.}
\centering
\footnotesize
\setlength{\tabcolsep}{2.5pt}
\renewcommand{\arraystretch}{1.08}
\begin{tabular}{ccccccccccccc}
\midrule[1.2pt]
\multirow{2}{*}{Method}                  & \multicolumn{3}{c}{LiveBench}                 & \multicolumn{3}{c}{MBPP}                      & \multicolumn{3}{c}{CodeContests}              & \multicolumn{3}{c}{CodeForces}                \\ \cline{2-13} 
                                         & Pass@1        & Pass@5        & APR           & Pass@1        & Pass@5        & APR           & Pass@1        & Pass@5        & APR           & Pass@1        & Pass@5        & APR           \\ \hline
MapCoder                                 & 32.7          & 35.9          & 45.2          & 72.3          & 75.7          & 80.4          & 22.8          & 29.5          & 40.7          & 7.4           & 8.4           & 13.7          \\
\alg$_{\text{Retrieval}}$ & \textbf{36.7} & \textbf{38.9} & 45.8          & \textbf{74.4} & \textbf{77.4} & \textbf{82.5} & 23.7          & 31.4          & 40.9          & 9.4           & 10.5          & \textbf{14.2} \\
\alg$_{\text{Debugging}}$ & 35.4          & 37.8          & \textbf{46.2} & 73.8          & 76.7          & 81.9          & \textbf{24.4} & \textbf{33.7} & \textbf{42.7} & \textbf{10.2} & \textbf{11.2} & \textbf{14.2} \\ \midrule[1.2pt]
\end{tabular}
\label{tab:agent_rl_other}
\end{table}

Table~\ref{tab:agent_rl_other} shows that \alg can be effectively extended beyond the Planner--Coder paradigm to other multi-agent code generation frameworks. When applied to MapCoder, independently reinforcing either the Retrieval Agent or the Debugging Agent consistently improves performance over the original MapCoder baseline across all benchmarks. Specifically, \alg$_{\text{Retrieval}}$ achieves stronger gains on LiveBench and MBPP, suggesting that optimizing the retrieval process helps provide more useful exemplars and algorithmic hints for downstream code generation. In contrast, \alg$_{\text{Debugging}}$ obtains the best results on CodeContests and CodeForces, indicating that reinforcement of the debugging process is particularly beneficial for more challenging competitive-programming tasks where execution feedback and iterative correction are crucial. Overall, these results demonstrate that \alg is not limited to a specific Planner--Coder architecture, but can serve as a general collaboration-aware reinforcement learning strategy for improving different agents within broader multi-agent systems.


\subsection{Sensitivity Analysis}

\begin{figure*}
    \centering
    \includegraphics[width=0.95\linewidth]{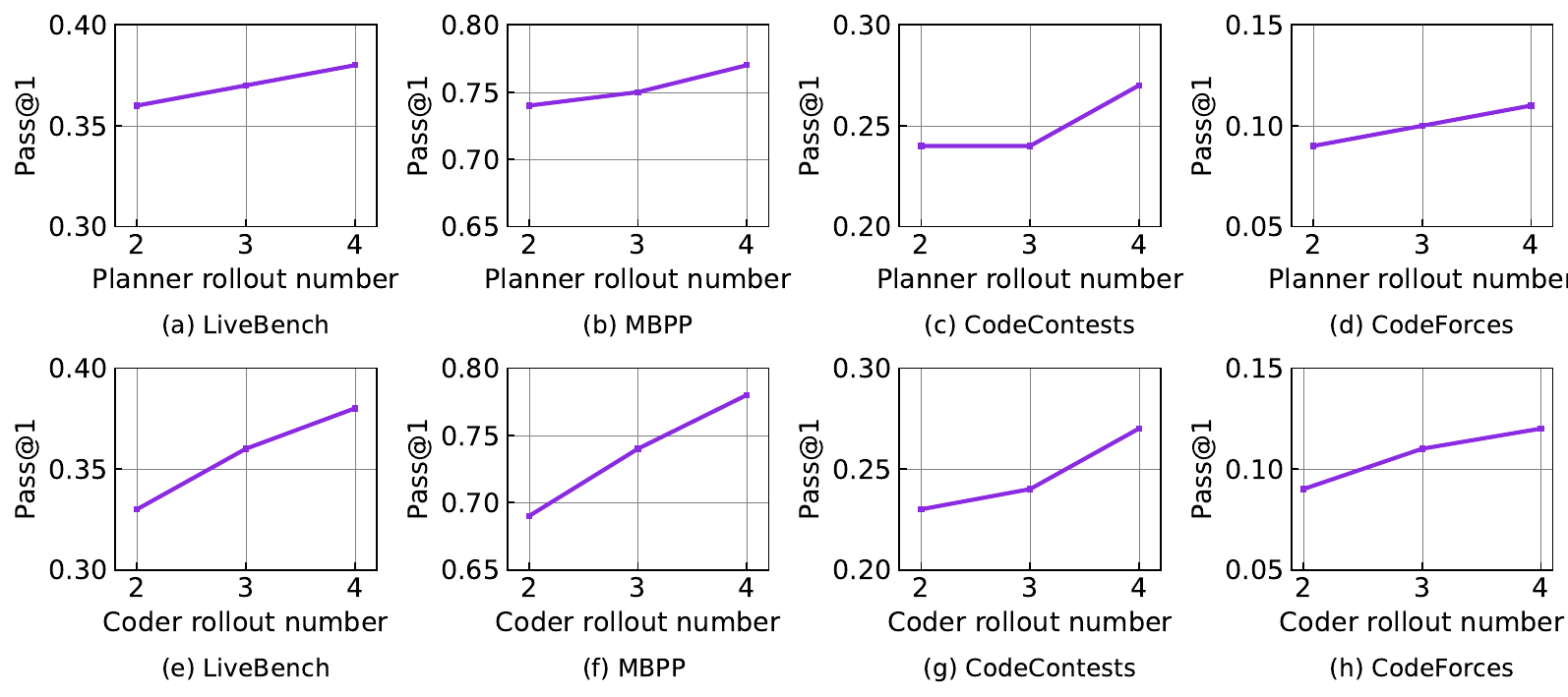}
    \caption{Impact of RL's number of rollouts for Planner and Coder agent. Qwen2.5‑7B‑Coder‑Instruct is considered as base model.}
    \label{fig:rollout}
\end{figure*}

Figure~\ref{fig:rollout} shows the sensitivity of GRPO training to the rollout number for both the Planner and Coder agents. We observe that increasing the rollout number from 2 to 4 consistently improves Pass@1 across all benchmarks, suggesting that a larger rollout budget provides richer exploration and more reliable optimization signals. For the Planner agent, larger rollouts help discover better high-level solution strategies, while for the Coder agent, they improve the generation of executable implementations conditioned on the plan. The improvement is especially clear for the Coder agent on LiveBench and MBPP, indicating that code generation is more sensitive to rollout diversity. Overall, these results show that \alg remains robust under different rollout settings and generally benefits from using more GRPO rollouts.

\section{Conclusion}
\label{sec:conc}

We propose \alg, a collaborative framework that combines cold‑start specialization and reinforcement learning to enable a Planner and a Coder agent to work together for plan‑to‑code translation. Experiments across multiple benchmarks show that \alg consistently improves accuracy, efficiency, and maintainability over base models and existing methods, demonstrating the effectiveness of role‑specialized collaboration for LLM‑based code generation.

\bibliography{reference}
\bibliographystyle{unsrtnat}

\appendix

\section{Limitations}

Although CoRe-Code achieves promising results, this work still has several minor limitations. 
First, our experiments are conducted on a selected set of representative code generation benchmarks and open-source base models, and more model families can be explored in future work. 
Second, we mainly follow a fixed Planner--Coder interaction format, while alternative prompting templates or more diverse agent communication styles may further improve performance. 

\section{Broader Impacts}

CoRe-Code aims to improve the accuracy and efficiency of LLM-based code generation by enhancing collaboration between specialized agents. 
This work may benefit developers, researchers, and students by reducing the effort required to generate correct and efficient code, especially for algorithmic programming tasks. 
By encouraging the Planner agent to produce more structured algorithmic thoughts and the Coder agent to better follow them, CoRe-Code may also improve the interpretability of intermediate reasoning in code generation systems.

At the same time, more capable code generation systems may be misused to generate incorrect, insecure, or harmful code if deployed without proper validation. 
Therefore, generated programs should still be checked through testing, human review, and security analysis before being used in real-world software systems. 
Our experiments focus on standard code generation benchmarks and open-source models, and the proposed framework is intended to support responsible and verifiable code generation research.

\section{Time-Complexity Prediction}
\label{sec:complex}

\begin{algorithm}[H] \caption{Time-Complexity Prediction for Planner Optimization.} \label{algo:time_complexity_prediction} \begin{algorithmic}[1] \renewcommand{\algorithmicrequire}{\textbf{Input:}} \renewcommand{\algorithmicensure}{\textbf{Output:}} \Require Problem instance $x_i$, planner-generated plan $p_i$, coder-generated code $c_i$, complexity class set $\mathcal{C}$, complexity predictor $\mathcal{M}_{\phi}$. \Ensure Predicted time complexity $\hat{\tau}_i$ and time-complexity reward $r_{\text{time}_i}$. \State $\mathcal{F}_i \leftarrow \emptyset$. \State \textcolor{blue}{// Step I: Program-Plan Structural Analysis.} \State Parse the generated code $c_i$ into an abstract syntax tree $\mathcal{A}_i$. \State Extract input-size variables $\bm{n}_i$ from the problem instance $x_i$ and code $c_i$. \State Extract structural features $\mathcal{F}_i$ from $\mathcal{A}_i$ and planner-generated plan $p_i$. \label{step_time_one} \State \textcolor{blue}{// Step II: Time-Complexity Class Prediction.} \State Estimate the complexity distribution: \[ \bm{q}_i \leftarrow \mathcal{M}_{\phi}(x_i, p_i, c_i, \mathcal{F}_i). \] \State Predict the time-complexity class: \[ \hat{\tau}_i \leftarrow \arg\max_{\tau \in \mathcal{C}} \bm{q}_i(\tau). \] \State Compute the prediction confidence: \[ \rho_i \leftarrow \max_{\tau \in \mathcal{C}} \bm{q}_i(\tau). \] \label{step_time_two} \State \textcolor{blue}{// Step III: Time-Complexity Reward Construction.} \State Map the predicted complexity class $\hat{\tau}_i$ to a normalized cost score: \[ s_i \leftarrow \operatorname{Cost}(\hat{\tau}_i). \] \State Compute the time-complexity reward: \[ r_{\text{time}_i} \leftarrow \rho_i \cdot (1 - s_i). \] \label{step_time_three} \State \Return $\hat{\tau}_i$, $r_{\text{time}_i}$. \end{algorithmic} \end{algorithm}

\section{Related Work and Background}
\label{rw:supp}

\subsection{Enhancing Code Generation with RL and Multi-Agent Systems}

\myparatight{RL-based enhancement}
LLMs acquire foundational programming knowledge during pre-training, and their ability to follow instructions is further enhanced through SFT \citep{zhang2025sealign}. However, prior studies have shown that the generalization ability of SFT is often limited, with models tending to overfit to training distributions and struggling on out‑of‑distribution tasks \citep{korbak2023pretraining}.
To further adapt these models to real-world deployment scenarios, RL serves as an effective technique, enabling them to excel in diverse and complex applications \citep{kumar2025llm}.

Reinforcement Learning from Human Feedback (RLHF) \citep{ouyang2022training}, originally designed for natural language generation, has seen growing application in code generation \citep{zhang2024policy,wang2024enhancing}, where it is used to guide models toward producing outputs that better adhere to developer intentions, coding conventions, and correctness requirements. However, RLHF requires training a reward model and using Proximal Policy Optimization (PPO), a powerful RL method, which can be unstable and resource-intensive. To mitigate this, Direct Preference Optimization (DPO) \citep{rafailov2023direct} offers a simpler, more stable alternative that directly learns from human preferences without explicit reward modeling or RL.
The DPO and its variants have also \citep{chen2024noise} demonstrated promising results in code generation. Chen et al. \citep{chen2024noise,zhang2025focused} introduced InfoNCA, an alignment framework that unifies the processing of explicit reward and preference data, extending the coding capabilities of DPO. Zhang et al \citep{zhang2025focused} propose Focused-DPO, which introduces fine-grained identification and optimization of error-prone points in code. By restructuring the reward function, it emphasizes these critical segments with increased weight during training.

Recently, DeepSeek R1 \citep{guo2025deepseek} has demonstrated impressive reasoning capabilities on complex tasks, particularly excelling in code generation. Its RL algorithm, Gradient-based Reinforcement Policy Optimization (GRPO) \citep{shao2024deepseekmath}, exhibits strong generalization and significant performance gains in complex code reasoning and generation tasks. In section \ref{sec:grpo}, we provide a detailed introduction to it.

\myparatight{Multi-Agent Systems-based enhancement} Compared to a multi-agent system, a single LLM that generates code directly or uses pseudo-code approaches like CoT often struggles to produce complete solutions for complex problems \citep{islam2024mapcoder}. Multi-agent systems enable more flexible, efficient, and interpretable task-solving through role specialization, collaborative reasoning, and tool integration \citep{huang2023agentcoder}. Huang et al. \citep{huang2023towards} proposed a test executor agent that leverages a Python interpreter to generate test logs for LLMs. Similarly, Zhong et al. \citep{zhong2024debug} introduced a debugger agent that employs a static analysis tool to construct control flow graphs, helping LLMs identify bug locations more effectively. Islam et al. \citep{islam2024mapcoder} propose MapCoder, a multi-agent framework that mimics the human coding process through retrieval, planning, coding, and debugging. It achieves state-of-the-art results on diverse programming benchmarks, showing strong generalization and robustness on complex tasks.
Lin et al. \citep{li2025structured} introduces FlowGen, a multi-agent framework that simulates software process models with role-based LLMs, achieving superior code quality and stability over baselines through structured collaboration.

\subsection{Preliminaries on GRPO}
\label{sec:grpo}

Group Relative Policy Optimization (GRPO) \citep{shao2024deepseekmath} is a modified version of PPO that optimizes policies using policy gradients derived from reward-based losses. It encourages the exploration of richer and more diverse reasoning paths by comparing responses sampled within the same group.

Formally, let $Q$ be the question set, which contains various programming questions along with their accompanying test cases.
Let $\pi_{\theta_{\text{old}}}$ be the current policy model, and $\{o_1, o_2, \dots, o_G\}$ a group of responses generated by $\pi_{\theta_{\text{old}}}$ for question $q$. Let $\pi_{\theta_{\text{ref}}}$ denote the frozen reference model. The GRPO optimization objective is defined as follows:
{\footnotesize \begin{equation}
\begin{aligned}
&J_{\text{GRPO}}(\theta) = \mathbb{E}_{q \sim Q, \{o_i\}_{i=1}^G \sim \pi_{\theta_{\text{old}}}} \\
&\left[
\frac{1}{G} \sum_{i=1}^G \sum_{t=1}^{|o_i|} 
\min\left(
\frac{\pi_{\theta}(o_{i,t}|q)}{\pi_{\theta_{\text{old}}}(o_{i,t}|q)} A_i,\,
\text{clip}\left(
\frac{\pi_{\theta}(o_{i,t}|q)}{\pi_{\theta_{\text{old}}}(o_{i,t}|q)},
1-\epsilon, 1+\epsilon
\right) A_i
\right)
- \beta D_{\text{KL}}(\pi_\theta \| \pi_{\text{ref}})
\right]
\label{grpo}
\end{aligned}
\end{equation} }
Here, $\epsilon$ and $\beta$ denote the clipping threshold and the coefficient for the KL-divergence penalty, respectively. The advantage $A_i$ for each response is calculated as:
{\small \begin{equation}
A_i = \frac{r_i - \text{mean}(\{r_1, r_2, \dots, r_G\})}{\text{std}(\{r_1, r_2, \dots, r_G\})}, \quad \text{where} \, \{r_i\}_{i=1}^{G} \, \text{are reward set.}
\label{advantages}
\end{equation} }
GRPO replaces the critic model used in PPO with a more efficient intra-group advantage estimation, reducing computational overhead.

\section{Extra experiment}

\subsection{Experiment details}
\label{sec:details}

In this experiment, both agents were trained with reinforcement learning under the same hyperparameter settings. Specifically, the learning rate (\texttt{lr}) was set to $1.0 \times 10^{-6}$, the weight decay (\texttt{weight\_decay}) was set to $1.0 \times 10^{-2}$, and the optimizer was \texttt{adamw} (with \texttt{adamw\_bf16} as an alternative option). The learning-rate warmup ratio (\texttt{lr\_warmup\_ratio}) was set to 0, and the number of rollout samples was fixed to 5. During reinforcement learning, the Planner Agent was trained for 100 steps on Qwen2.5-Coder-7B and Qwen2.5-Coder-14B, and for 50 steps on Qwen3-4B, while the Coder Agent was trained for 150 steps across all models.

The reinforcement learning stage used 15,000 training samples from \citep{xu2025kodcode,li2023taco}. All reinforcement learning baselines and our method were trained using the same data split. We further verified that there was no data leakage between the training and evaluation sets.



\subsection{Extra experiment results}
\label{sec:extra_exp}

\subsubsection{Efficiency and maintainability of the generated code}

\begin{table*}[h]
\footnotesize
\centering
\caption{The efficiency and maintainability of the generated code are evaluated using Runtime (\(\downarrow\)), MU (\(\downarrow\)), and CC (\(\downarrow\)), where lower values of Runtime, MU, and CC indicate better performance. \textbf{Bold} indicates the best performance for clarity.
}
\begin{tabular}{ccccccccccccc}
\midrule[1.2pt]
\multirow{2}{*}{Method}                & \multicolumn{3}{c}{LiveBench}                & \multicolumn{3}{c}{MBPP}                     & \multicolumn{3}{c}{CodeContests}             & \multicolumn{3}{c}{CodeForces}               \\ \cline{2-13} 
                                       & Runtime         & MU          & CC           & Runtime         & MU          & CC           & Runtime         & MU          & CC           & Runtime         & MU          & CC           \\ \hline
Qwen2.5-7B-Instruct                    & 7.5          & 525.3          & 4.4          & 6.9          & 229.6          & 2.1          & 6.9          & 625.4          & 5.4          & \textbf{2.9} & 754.7          & \textbf{6.9}          \\
\cellcolor{greyL} \alg$_{\text{w/ 7B-Instruct}}$ & \cellcolor{greyL}\textbf{6.4} & \cellcolor{greyL}\textbf{452.7} & \cellcolor{greyL}\textbf{3.9} & \cellcolor{greyL}\textbf{6.2} & \cellcolor{greyL}\textbf{198.4} & \cellcolor{greyL}\textbf{1.6} & \cellcolor{greyL}\textbf{6.0} & \cellcolor{greyL}\textbf{600.4} & \cellcolor{greyL}\textbf{5.1} & \cellcolor{greyL}3.5          & \cellcolor{greyL}\textbf{713.4} & \cellcolor{greyL}7.2 \\ \hline
Qwen2.5-7B-Coder-Instruct              & 7.1          & 501.4          & 4.1          & 6.6          & 208.4          & 1.8          & 6.7          & \textbf{603.6} & 5.2          & 2.8          & 747.1          & 6.8          \\
\cellcolor{greyL}\alg$_{\text{w/ 7B-Coder}}$    & \cellcolor{greyL}\textbf{6.6} & \cellcolor{greyL}\textbf{473.2} & \cellcolor{greyL}\textbf{3.6} & \cellcolor{greyL}\textbf{6.3} & \cellcolor{greyL}\textbf{197.7} & \cellcolor{greyL}\textbf{1.7} & \cellcolor{greyL}\textbf{6.2} & \cellcolor{greyL}611.7          & \cellcolor{greyL}\textbf{4.8} & \cellcolor{greyL}\textbf{2.6} & \cellcolor{greyL}\textbf{719.7} & \cellcolor{greyL}\textbf{6.4} \\ \midrule[1.2pt]
\end{tabular}
\label{Tab:effiency}
\end{table*}
We evaluate the efficiency and maintainability of code generated by \alg. 
The average runtime and average MU are used to measure time complexity and space complexity, respectively, while cyclomatic complexity (CC) is adopted as the maintainability metric. 
The base models are Qwen2.5-7B-Instruct and Qwen2.5-7B-Coder-Instruct, denoted as \alg$_{\text{w/ 7B-Instruct}}$ and \alg$_{\text{w/ 7B-Coder}}$, and the results are reported in Table~\ref{Tab:effiency}.

Table~\ref{Tab:effiency} evaluates the efficiency and maintainability of the generated code in terms of Runtime, memory usage (MU), and cyclomatic complexity (CC). Overall, \alg improves most metrics across different benchmarks and base models, showing that the proposed collaboration-aware training does not merely improve functional correctness, but also encourages the generation of more efficient and maintainable code. Compared with Qwen2.5-7B-Instruct, \alg reduces Runtime, MU, and CC on LiveBench, MBPP, and CodeContests, with particularly clear reductions in memory usage. Similar improvements can be observed when using Qwen2.5-7B-Coder-Instruct as the base model, where \alg achieves lower Runtime and CC on all four benchmarks, and lower MU on most datasets. Although there are a few minor regressions, such as Runtime and CC on CodeForces for the general 7B model and MU on CodeContests for the coder model, the overall trend demonstrates that \alg can improve code quality beyond accuracy by producing solutions with better execution efficiency and lower structural complexity.

\subsubsection{Reliability of generated code}
To examine whether the enhanced algorithmic thought planning in \alg can mitigate code generation errors, we adopt the failure rate (FR) as the evaluation metric \citep{wang2025towards}. To further refine the analysis, we focus on the proportions of the three most common error types, namely timeout errors (TOE), value errors (VE), and type errors (TE), across all generated samples. We selected Qwen2.5‑7B‑Instruct and Qwen2.5‑7B‑Coder‑Instruct as the base models for \alg. In Table~\ref{Tab:effiency}, they are denoted as \alg${\text{w/ 7B‑Instruct}}$ and \alg${\text{w/ 7B‑Coder}}$, respectively.

\begin{table*}[h]
\tiny
\centering
\setlength\tabcolsep{4pt}
\caption{The performance of the generated code is evaluated using FR (\(\downarrow\)), TOE (\(\downarrow\)), VE (\(\downarrow\)), and TE (\(\downarrow\)), where lower values of FR, TOE, VE, and TE indicate better performance. \textbf{Bold} highlights the best performance for clarity.
}
\scriptsize
\begin{tabular}{ccccccccccccccccc}
\midrule[1.2pt]
\multirow{2}{*}{Method}                & \multicolumn{4}{c}{LiveBench}                              & \multicolumn{4}{c}{MBPP}                                    & \multicolumn{4}{c}{CodeContests}                           & \multicolumn{4}{c}{CodeForces}                              \\ \cline{2-17} 
                                       & FR            & TOE          & VE           & TE           & FR           & TOE           & VE           & TE            & FR            & TOE          & VE           & TE           & FR            & TOE           & VE           & TE           \\ \hline
Qwen2.5-7B-Instruct                    & 25.7          & 6.9          & 2.7          & 1.7          & 8.2          & \textbf{1.3} & 2.0          & \textbf{0.4} & 22.9          & 3.1          & 1.3          & 3.9          & 23.3          & 5.8           & 8.1          & \textbf{2.7} \\
\cellcolor{greyL}\alg$_{\text{w/ 7B-Instruct}}$ & \cellcolor{greyL}\textbf{16.7} & \cellcolor{greyL}\textbf{2.3} & \cellcolor{greyL}\textbf{1.2} & \cellcolor{greyL}\textbf{1.0} & \cellcolor{greyL}\textbf{4.0} & \cellcolor{greyL}1.9           & \cellcolor{greyL}\textbf{0.8} & \cellcolor{greyL}0.6           & \cellcolor{greyL}\textbf{13.2} & \cellcolor{greyL}\textbf{2.5} & \cellcolor{greyL}\textbf{0.9} & \cellcolor{greyL}\textbf{1.4}         & \cellcolor{greyL}\textbf{20.3} & \cellcolor{greyL}\textbf{2.6}  & \cellcolor{greyL}\textbf{4.7} & \cellcolor{greyL}3.5          \\ \hline
Qwen2.5-7B-Coder-Instruct              & 28.4          & \textbf{3.8} & 3.7          & 1.8          & 6.2          & \textbf{1.7}  & \textbf{0.9} & 0.3           & 25.5          & 3.1          & 1.9          & \textbf{1.2} & 30.6          & 11.8          & \textbf{6.1} & 2.1 \\
\cellcolor{greyL}\alg$_{\text{w/ 7B-Coder}}$    & \cellcolor{greyL}\textbf{15.7} & \cellcolor{greyL}5.4          & \cellcolor{greyL}\textbf{2.2} & \cellcolor{greyL}\textbf{1.2} & \cellcolor{greyL}\textbf{5.9} & \cellcolor{greyL}2.3           & \cellcolor{greyL}1.2          & \cellcolor{greyL}\textbf{0.2}  & \cellcolor{greyL}\textbf{15.4} & \cellcolor{greyL}\textbf{2.5} & \cellcolor{greyL}\textbf{1.5} & \cellcolor{greyL}1.9          & \cellcolor{greyL}\textbf{23.8} & \cellcolor{greyL}\textbf{9.4} & \cellcolor{greyL}7.0          & \cellcolor{greyL}\textbf{1.4}          \\ \midrule[1.2pt]
\end{tabular}
\label{Tab:error}
\end{table*}

Table~\ref{Tab:error} further analyzes the error characteristics of the generated code using failure rate (FR), timeout error (TOE), value error (VE), and type error (TE). Overall, \alg substantially reduces the occurrence of execution failures across different benchmarks and base models, indicating that the proposed collaboration-aware training improves not only the final pass rate but also the robustness of generated programs. The most consistent improvement is observed on FR: for both Qwen2.5-7B-Instruct and Qwen2.5-7B-Coder-Instruct, \alg lowers the failure rate on all four benchmarks, with particularly large reductions on LiveBench and CodeContests. This suggests that better coordination between the Planner and Coder helps generate solutions that are less likely to fail during execution.

In addition, \alg generally reduces TOE and VE, showing that the generated programs become less prone to inefficient execution and incorrect intermediate computation. For example, with the 7B-Instruct base model, \alg achieves lower TOE and VE on LiveBench, CodeContests, and CodeForces. With the 7B-Coder base model, \alg also reduces TOE on CodeContests and CodeForces, and reduces VE on LiveBench and CodeContests. Although a few metrics slightly increase, such as TOE on MBPP and TE on CodeForces for the 7B-Instruct model, the overall trend shows that \alg effectively suppresses common execution errors. These results demonstrate that \alg improves the reliability and stability of generated code by reducing both overall failures and specific error types.

\subsubsection{Runtime Comparison among Multi-Agent Systems}

\begin{figure*}[t]
    \centering

    \subfloat[LiveBench]{\includegraphics[width=0.23\textwidth]{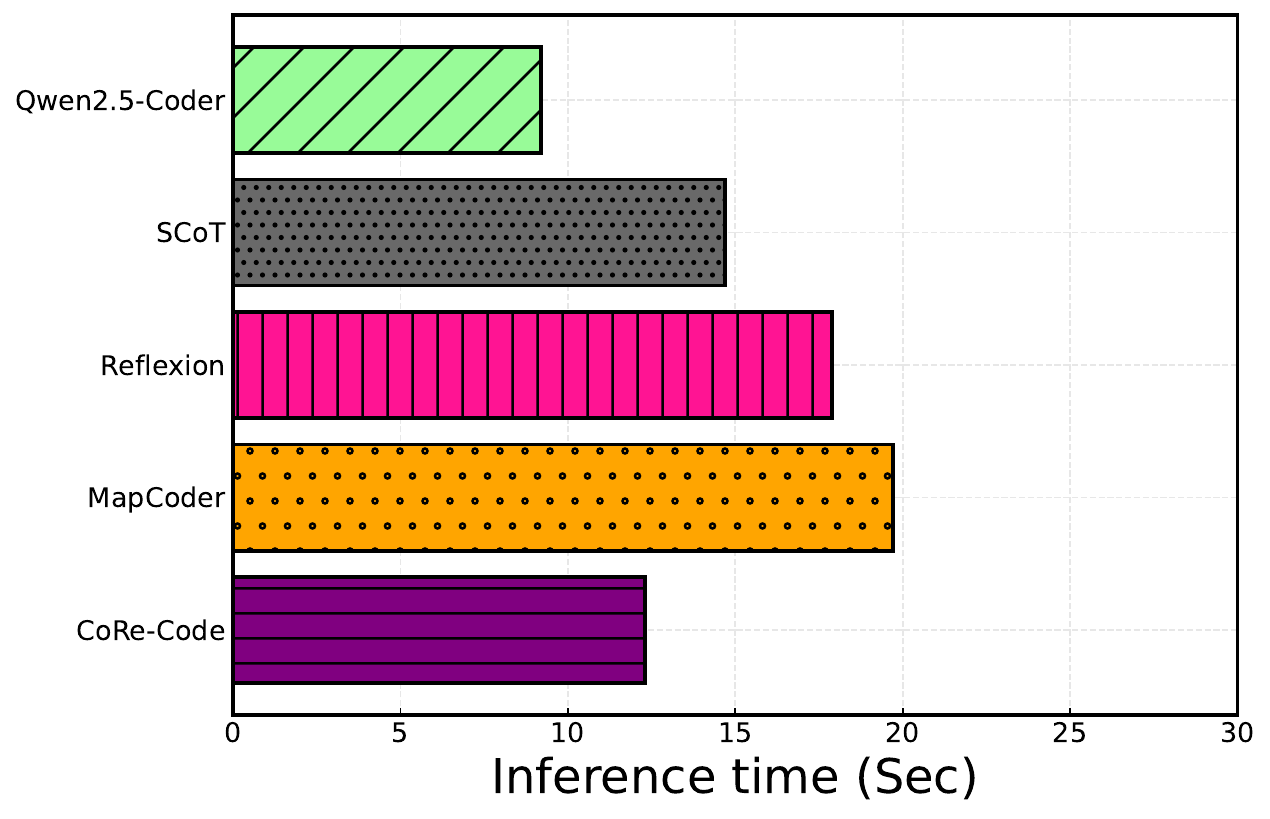}\label{fig:MNIST_time}}
    \subfloat[MBPP]{\includegraphics[width=0.23\textwidth]{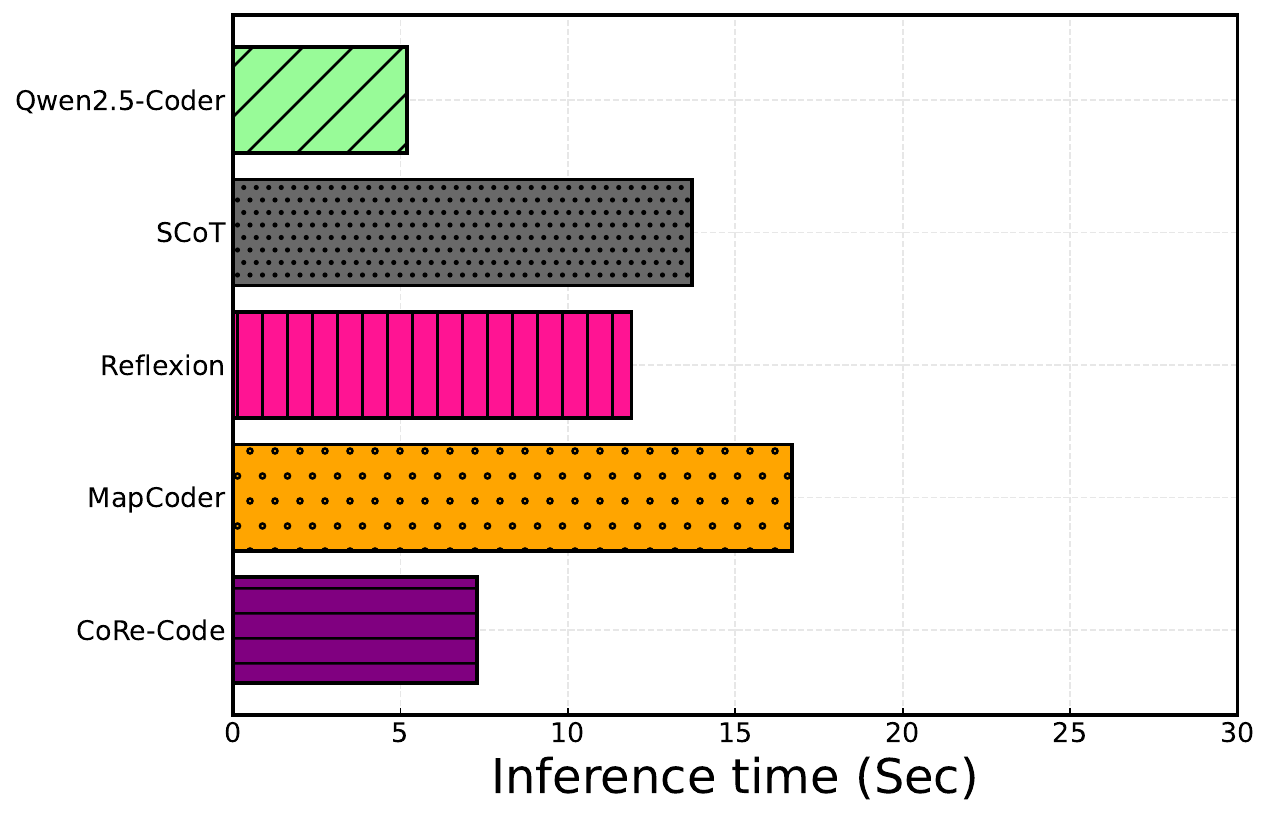}\label{fig:FEMNIST_time}}
    \subfloat[CodeContests]{\includegraphics[width=0.23\textwidth]{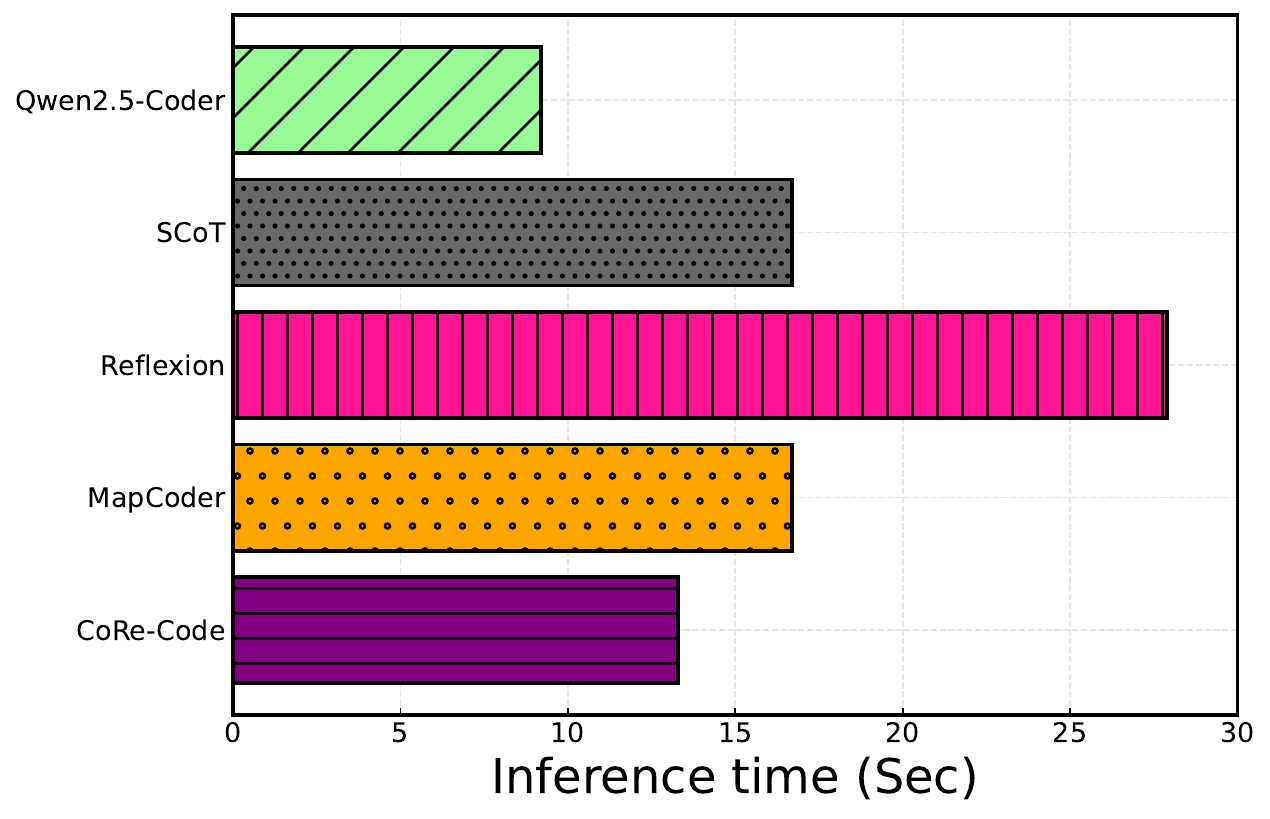}\label{fig:STL10_time}}
    \subfloat[CodeForces]{\includegraphics[width=0.23\textwidth]{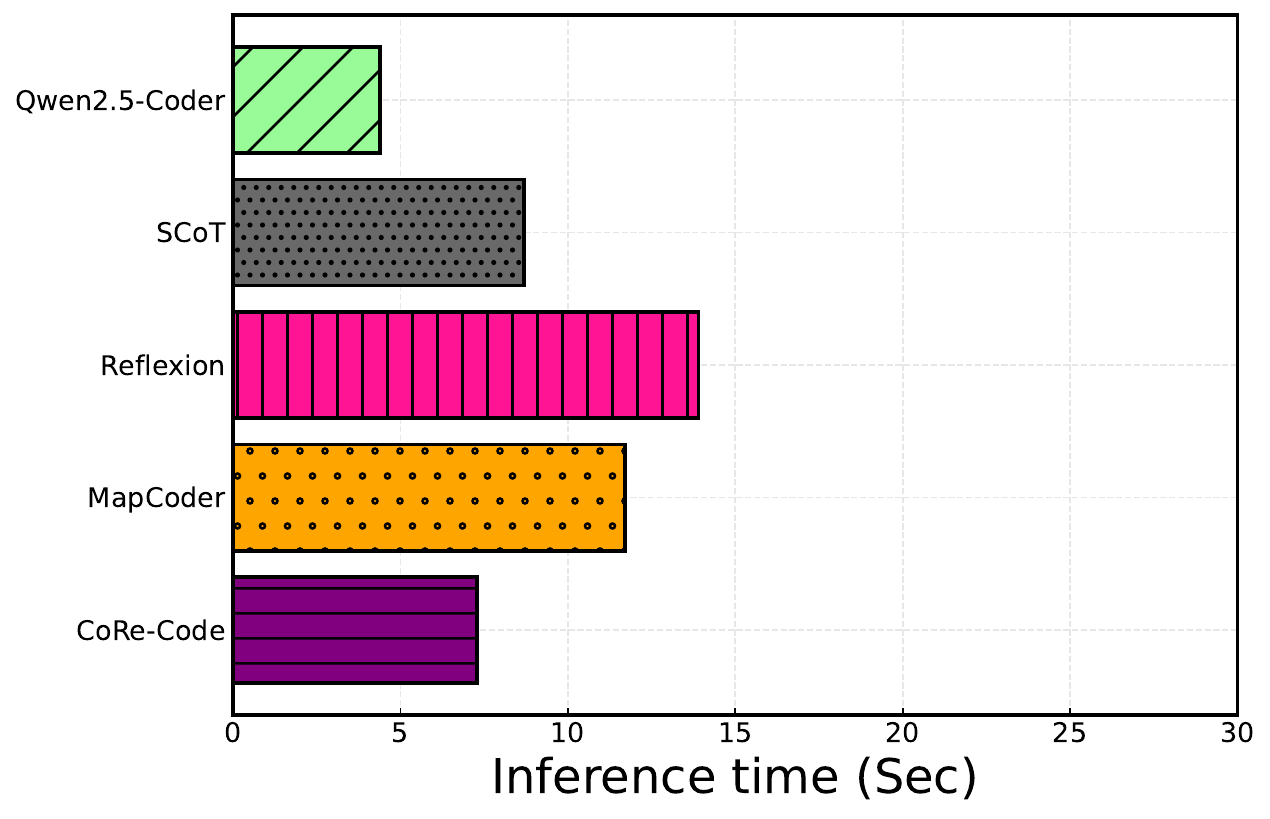}\label{fig:CIFAR10_time}}

    \caption{Computation costs of different methods, where Qwen2.5‑7B‑Coder‑Instruct is considered as base model.}
    \label{fig:time_comsume_fig}
\end{figure*}

Figure~\ref{fig:time_comsume_fig} compares the inference-time cost of different multi-agent code generation methods across four benchmarks. As expected, the single-agent Qwen2.5-Coder baseline has the lowest runtime because it does not involve additional agent interaction or iterative refinement. Among multi-agent methods, Reflexion and MapCoder usually introduce larger computational overhead due to repeated self-reflection, retrieval, planning, debugging, or multi-step communication. In contrast, CoRe-Code achieves a more favorable efficiency--performance trade-off: although it requires extra computation compared with the single-agent baseline, its inference time is consistently lower than or comparable to most multi-agent baselines, especially Reflexion and MapCoder. This suggests that the Planner--Coder collaboration in CoRe-Code improves code generation without relying on excessively long inference chains. Overall, the results show that CoRe-Code maintains the benefit of multi-agent collaboration while keeping the computational cost relatively controlled.

\newpage

\subsection{Training dynamics of \alg}
\label{sec:test_case}

\begin{figure}[H]
    \centering
    \subfloat[Accuracy reward]{
        \includegraphics[width=0.45\linewidth]{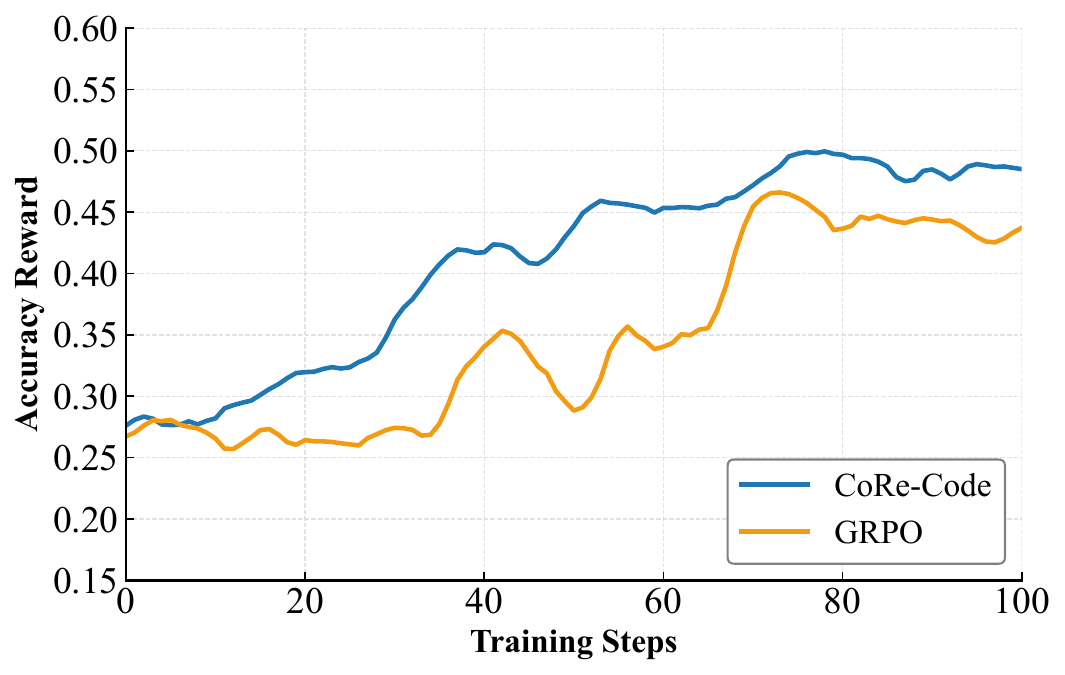}
        \label{fig:code_accuracy}
    }
    \hfill
    \subfloat[Entropy loss]{
        \includegraphics[width=0.45\linewidth]{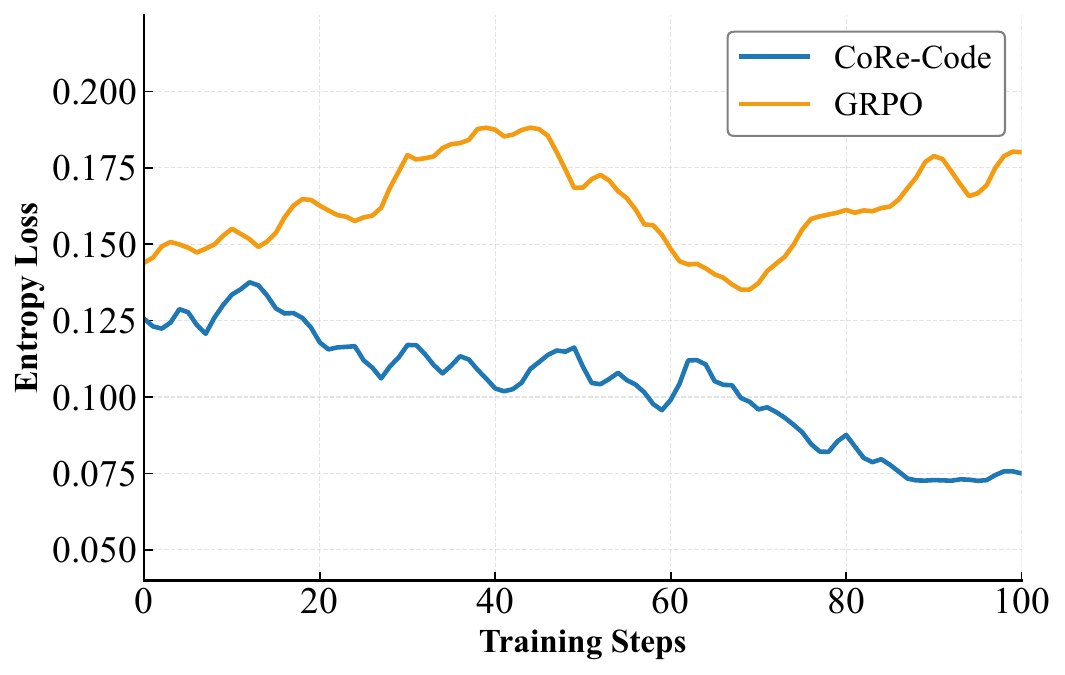}
        \label{fig:code_entropy}
    }
    \caption{Training dynamics of \alg, where Qwen2.5-7B-Coder-Instruct is used as the base model.}
    \label{fig:dynamic}
\end{figure}

During the reinforcement learning process, we present the training dynamics in Figure \ref{fig:dynamic}, illustrating the changes in accuracy, reward, and entropy, and compare our approach against GRPO. The results demonstrate that our method consistently achieves higher accuracy and reward, while maintaining lower entropy. This indicates that under the global guidance of the planner, the coder agent is able to acquire more valuable strategies more efficiently, exhibiting greater stability and confidence in decision-making with reduced reliance on high-randomness exploration. Therefore, the planner–coder framework not only enhances task performance but also encourages the agent to develop more deterministic policy choices, leading to simultaneous improvements in both efficiency and effectiveness.

\subsection{Example of \alg}
\label{sec:test_case}

\begin{promptbox}{\alg (Sample 1)}
Q: Write a function to replace all spaces in the given string with character * list item * list item * list item * list item '
\begin{lstlisting}[style=plantext]
<plan>
Input:
- s (string): The original input string which may contain spaces.
- Output:
- A new string where every space character `' '` in s is replaced with the sequence `'%20'`.

Step 1: Define the Input-Output Structure

 Input: A single string `s`.
 Output: A transformed string in which all space characters are substituted with `'%20'`.


Step 2: Solving Logic

 Sequence:

  1. Initialize an empty result container (conceptually a list of characters or substrings).
  2. Traverse the input string from the first to the last character.

 Branch:

  1. For each character:

      If the character is a space (`' '`), append `'%20'` to the result.
      Else, append the character itself unchanged.

 Loop:

  1. A loop is required to process each character in the input string sequentially until the end is reached.

 Final Step:

  1. Combine all elements in the result container into a single string.
  2. Return this new string.
</plan>
\end{lstlisting}

\begin{lstlisting}[style=mypython]
def replaceSpaces(s):
    result = []
    for char in s:
        if char == ' ':
            result.append("%20")
        else:
            result.append(char)
    return "".join(result)
\end{lstlisting}
\end{promptbox}

\begin{promptbox}{\alg (Sample 2)}
Q: Write a function to get a lucid number smaller than or equal to n.
\begin{lstlisting}[style=plantext]
<plan>
Input:
- n (an integer): The upper bound for searching lucid numbers, where n >= 1.

Output:

The largest lucid number less than or equal to n.

1. Define Lucid Number:

   A lucid number is a positive integer whose prime factors are only 2, 3, 5, or 7.
   Examples: 1, 2, 3, 5, 7, 6, 10, 14, 15, 21, 28, etc.

2. Sequence (Step-by-Step Process):

   1. Start with the base set {1}, since 1 is trivially lucid.
   2. Generate new lucid numbers by multiplying existing ones by 2, 3, 5, or 7.
   3. Continue generating until values exceed n.
   4. Store all generated lucid numbers in a data structure (e.g., list or heap).
   5. Select the maximum lucid number that is <= n.

3. Branch (Conditions):

   If n < 1 -> no lucid number exists, return 0 or handle as invalid input.
   If n itself is lucid -> directly return n.
   Otherwise, search among generated lucid numbers for the largest <= n.

4. Loop (Repetition):

   Use a loop or priority-based expansion to multiply numbers repeatedly until exceeding n.
   Continue until no new lucid numbers can be generated under the bound.
</plan>
\end{lstlisting}

\begin{lstlisting}[style=mypython]
def largest_lucid_number(n):
    if n < 1:
        return 0
    primes = [2, 3, 5, 7]
    lucid = {1}
    queue = [1]

    while queue:
        current = queue.pop()
        for p in primes:
            new_val = current * p
            if new_val <= n and new_val not in lucid:
                lucid.add(new_val)
                queue.append(new_val)

    return max(x for x in lucid if x <= n)
\end{lstlisting}
\end{promptbox}

\begin{promptbox}{\alg (Sample 3)}
Q: 
Due to the recent popularity of the Deep learning new countries are starting to look like Neural Networks. That is, the countries are being built deep with many layers, each layer possibly having many cities. They also have one entry, and one exit point.

There are exactly L layers, each having N cities. Let us look at the two adjacent layers L1 and L2. Each city from the layer L1 is connected to each city from the layer L2 with the traveling cost cij for $$i,j\in\{1,2,\ldots,N\}$$, and each pair of adjacent layers has the same cost in between their cities as any other pair (they just stacked the same layers, as usual). Also, the traveling costs to each city from the layer L2 are same for all cities in the L1, that is cij is the same for $$i \in \{1, 2, \ldots, N\}$$, and fixed j.

Doctor G. needs to speed up his computations for this country so he asks you to find the number of paths he can take from entry to exit point such that his traveling cost is divisible by given number M.
\begin{lstlisting}[style=plantext]
<plan>
- L (integer): number of layers in the country (excluding entry/exit).
- N (integer): number of cities in each layer.
- c (list of integers of length N): traveling costs to each city in the next layer. For all i in layer L1, cost to city j in layer L2 is `c[j]`.
- M (integer): the modulus divisor.

Output:

 count (integer): the number of distinct paths from entry to exit such that the total traveling cost is divisible by M.

Strategy:

1. Input-Output Structure

    Input defines a stacked structure of L layers, each with N cities, connected in a complete bipartite way to the next layer with uniform costs per destination.
    Output is a single integer, representing the number of valid paths modulo or exact (depending on constraints).

2. Solving Logic

    Sequence:

     1. Model path costs as sequences of choices across L transitions (from entry to exit).
     2. Since costs depend only on the chosen city in each layer, the total cost is the sum of L chosen city costs (one from each layer).
     3. The number of paths equals the number of such sequences, which is $N^L$. But we only count sequences whose total cost % M == 0.
    Branch:

      For each possible remainder modulo M, decide whether adding a new city's cost keeps us on a path toward divisibility.
      Use a dynamic programming (DP) state:

        `dp[layer][r] = number of ways to reach remainder r after processing that many layers`.
    Loop:

      For each layer from 1 to L:

        For each remainder r:

          For each city j in that layer (N choices):

            Update next remainder: `(r + c[j]) % M`.
            Accumulate counts accordingly.
</plan>
\end{lstlisting}

\begin{lstlisting}[style=mypython]
def count_paths(L, N, costs, M):
    # Precompute frequencies of costs modulo M
    freq = [0] * M
    for c in costs:
        freq[c % M] += 1

    # Initialize DP: only 0 remainder has one path (empty path)
    dp = [0] * M
    dp[0] = 1

    # Process each layer
    for _ in range(L):
        new_dp = [0] * M
        for r in range(M):
            if dp[r] == 0:
                continue
            for mod_val in range(M):
                if freq[mod_val] == 0:
                    continue
                new_r = (r + mod_val) % M
                new_dp[new_r] += dp[r] * freq[mod_val]
        dp = new_dp

    # Answer is number of ways with remainder 0
    return dp[0]
\end{lstlisting}
\end{promptbox}


\newpage
\section*{NeurIPS Paper Checklist}

\begin{enumerate}

\item {\bf Claims}
    \item[] Question: Do the main claims made in the abstract and introduction accurately reflect the paper's contributions and scope?
    \item[] Answer: \answerYes{} 
    \item[] Justification: Appendix discusses the broader impacts of this work, covering both potential benefits and misuse risks. 
While improved reasoning reliability may benefit robust LLM-based systems, stronger reasoning models may also produce convincing yet incorrect solutions or be adapted for harmful multi-step tasks.

\item {\bf Limitations}
    \item[] Question: Does the paper discuss the limitations of the work performed by the authors?
    \item[] Answer: \answerYes{} 
    \item[] Justification: We have limitation.

\item {\bf Theory assumptions and proofs}
    \item[] Question: For each theoretical result, does the paper provide the full set of assumptions and a complete (and correct) proof?
    \item[] Answer: \answerNA{} 
    \item[] Justification: No theorical analysis in this paper.

    \item {\bf Experimental result reproducibility}
    \item[] Question: Does the paper fully disclose all the information needed to reproduce the main experimental results of the paper to the extent that it affects the main claims and/or conclusions of the paper (regardless of whether the code and data are provided or not)?
    \item[] Answer: \answerYes{} 
    \item[] Justification: All information can be founded in Appendix

\item {\bf Open access to data and code}
    \item[] Question: Does the paper provide open access to the data and code, with sufficient instructions to faithfully reproduce the main experimental results, as described in supplemental material?
    \item[] Answer: \answerYes{} 
    \item[] Justification: All code we have provided, and all training dataset is open source.

\item {\bf Experimental setting/details}
    \item[] Question: Does the paper specify all the training and test details (e.g., data splits, hyperparameters, how they were chosen, type of optimizer) necessary to understand the results?
    \item[] Answer: \answerYes{} 
    \item[] Justification: All hypermeter can be founded in Appendix.

\item {\bf Experiment statistical significance}
    \item[] Question: Does the paper report error bars suitably and correctly defined or other appropriate information about the statistical significance of the experiments?
    \item[] Answer: \answerNo{} 
    \item[] Justification: In this area, we don't need to do this.

\item {\bf Experiments compute resources}
    \item[] Question: For each experiment, does the paper provide sufficient information on the computer resources (type of compute workers, memory, time of execution) needed to reproduce the experiments?
    \item[] Answer: \answerYes{} 
    \item[] Justification: Yes, we have provided all information.
    
\item {\bf Code of ethics}
    \item[] Question: Does the research conducted in the paper conform, in every respect, with the NeurIPS Code of Ethics \url{https://neurips.cc/public/EthicsGuidelines}?
    \item[] Answer: \answerYes{} 
    \item[] Justification: We have done.

\item {\bf Broader impacts}
    \item[] Question: Does the paper discuss both potential positive societal impacts and negative societal impacts of the work performed?
    \item[] Answer: \answerYes{} 
    \item[] Justification: Yes, we have done.
    
\item {\bf Safeguards}
    \item[] Question: Does the paper describe safeguards that have been put in place for responsible release of data or models that have a high risk for misuse (e.g., pre-trained language models, image generators, or scraped datasets)?
    \item[] Answer: \answerNA{} 
    \item[] Justification: It is not suitable for our topic.

\item {\bf Licenses for existing assets}
    \item[] Question: Are the creators or original owners of assets (e.g., code, data, models), used in the paper, properly credited and are the license and terms of use explicitly mentioned and properly respected?
    \item[] Answer: \answerNA{} 
    \item[] Justification: It is not suitable for us.

\item {\bf New assets}
    \item[] Question: Are new assets introduced in the paper well documented and is the documentation provided alongside the assets?
    \item[] Answer: \answerNA{} 
    \item[] Justification: We does not release new assets.

\item {\bf Crowdsourcing and research with human subjects}
    \item[] Question: For crowdsourcing experiments and research with human subjects, does the paper include the full text of instructions given to participants and screenshots, if applicable, as well as details about compensation (if any)? 
    \item[] Answer: \answerNA{} 
    \item[] Justification: The paper does not involve crowdsourcing nor research with human subjects.

\item {\bf Institutional review board (IRB) approvals or equivalent for research with human subjects}
    \item[] Question: Does the paper describe potential risks incurred by study participants, whether such risks were disclosed to the subjects, and whether Institutional Review Board (IRB) approvals (or an equivalent approval/review based on the requirements of your country or institution) were obtained?
    \item[] Answer: \answerNA{} 
    \item[] Justification: The paper does not involve crowdsourcing nor research with human subjects.

\item {\bf Declaration of LLM usage}
    \item[] Question: Does the paper describe the usage of LLMs if it is an important, original, or non-standard component of the core methods in this research? Note that if the LLM is used only for writing, editing, or formatting purposes and does \emph{not} impact the core methodology, scientific rigor, or originality of the research, declaration is not required.
    \item[] Answer: \answerNA{} 
    \item[] Justification:  The core method development in this research does not involve LLMs as any important, original, or non-standard components.

\end{enumerate}

\end{document}